\documentclass[11pt]{article}

\usepackage[final]{acl}

\usepackage{times}
\usepackage{adjustbox}
\usepackage{fancyvrb}
\usepackage{fvextra}
\usepackage{latexsym}
\usepackage{amsmath}
\usepackage{amssymb}
\usepackage{multirow}
\usepackage{bm}
\usepackage{booktabs}
\usepackage{tabularx}
\usepackage{makecell}
\usepackage[table]{xcolor}
\usepackage{pifont}
\usepackage{array}
\usepackage[most]{tcolorbox}

\usepackage{arydshln}

\definecolor{oursrow}{HTML}{EDF4FA}

\definecolor{tickblue}{HTML}{3568A8}
\definecolor{crossorange}{HTML}{B26A2E}
\definecolor{baselinerow}{HTML}{F2F2F2}

\definecolor{deltapos}{HTML}{1B8E3E}  
\definecolor{deltaneg}{HTML}{D95F02}  

\newcommand{\posdelta}[1]{\textcolor{deltapos}{(+ #1)}}
\newcommand{\negdelta}[1]{\textcolor{deltaneg}{(- #1)}}

\newcommand{\cmark}{\textcolor{tickblue}{\ding{51}}}
\newcommand{\xmark}{\textcolor{crossorange}{\ding{55}}}

\definecolor{errorred}{HTML}{C43B3B}
\newcommand{\traptext}[1]{\textcolor{crossorange}{\textbf{#1}}}
\newcommand{\groundtext}[1]{\textcolor{deltapos}{\textbf{#1}}}
\newcommand{\badoutcome}[1]{\textcolor{errorred}{\textbf{#1}}}
\newcommand{\goodoutcome}[1]{\textcolor{deltapos}{\textbf{#1}}}

\newcolumntype{L}{>{\raggedright\arraybackslash}X}
\newcolumntype{C}{>{\centering\arraybackslash}p{0.72cm}}

\newcolumntype{Y}{>{\centering\arraybackslash}X}
\usepackage[T1]{fontenc}

\usepackage[utf8]{inputenc}

\usepackage{microtype}

\usepackage{inconsolata}

\usepackage{graphicx}

\title{From Relevant Posts to Attributable Spans: Explicit Symptom Evidence for Depression Questionnaire Inference}

\author{Sumanth Manduru \\
  George Mason University \\
  \texttt{smanduru@gmu.edu} \\\And
  Carlotta Domeniconi \\
  George Mason University \\
  \texttt{cdomenic@gmu.edu} \\}

\begin{document}
\maketitle
\begin{abstract}
Longitudinal social-media histories offer an opportunity to assess depression, potentially supporting the inference of standardized depression questionnaire responses. However, this task requires evidence that is semantically relevant to a symptom and attributable to the user, while being sufficiently specific to support an individual questionnaire item. Existing approaches rely on semantic similarity to retrieve posts, which can introduce misleading or non-attributable evidence, particularly in multi-symptom posts and discussions of others’ experiences. We propose SREE (Symptom Retrieval and Evidence-Grounded InfErence), an evidence-grounded pipeline that extracts first-person and symptom-specific text spans, selects a representative subset using a Facility Location-inspired objective, and performs Evidence-Gated Inference over the selected evidence. We evaluate evidence retrieval on BDI-Sen and evidence-conditioned questionnaire inference on eRisk 2019–2021. Compared with state-of-the-art retrieval approaches, SREE achieves relative improvements of 70\% in post-level precision and 59\% in F1, while retaining at least 30\% fewer posts. On multi-symptom sentences, it also outperforms dense retrieval in precision and F1 even with twice the retrieval budget. Across all three eRisk collections, SREE achieves the highest exact item-level agreement while remaining competitive on ordinal and aggregate metrics. These results highlight the importance of localizing attributable evidence and controlling its composition before inferring depression questionnaire responses.

\end{abstract}

\noindent\textbf{Content Warning:} \textcolor{blue}{This paper contains depression-related language that may be distressing.}

\section{Introduction}

Depression affects millions worldwide, yet many individuals do not receive timely care because of gaps in mental-health services \cite{bruffaerts2018treatmentgap,who2023depression}. Users often disclose emotional distress, daily struggles, and symptom-related experiences through social-media posts \cite{bucci2019digital}, providing a complementary source of longitudinal linguistic signals for scalable and interpretable assessment. However, such posts are informal and heterogeneous, and a single narrative may express multiple depression symptoms simultaneously. Depression screening has increasingly shifted toward symptom-grounded assessment using standardized clinical questionnaires such as the Beck Depression Inventory-II (BDI-II), with CLEF eRisk framing the task as inferring 21 item-level responses from social-media histories \cite{nguyen-etal-2022-improving, losada2019overview, perez2022automatic}.

Recent methods address this task by retrieving posts \cite{perez2023semantic, wang2024explainable, ravenda2025llms} or summaries \cite{aragon2024delving} relevant to individual BDI-II items before predicting responses \cite{ravenda2025transforming,  xu2026social}. These approaches identify evidence through post--questionnaire semantic similarity, which does not necessarily imply evidence about the user. A post may discuss a symptom through another person's experience, advice, quotations, or general statements, while still receiving high similarity to that symptom. Conversely, a long post may contain several symptoms, with weaker symptom cues diluted by unrelated or dominant content in a dense representation. These limitations undermine the utility of post-level similarity as a  proxy for attributable symptom evidence.

\begin{figure*}[t]
    \centering
      \includegraphics[width=\textwidth]{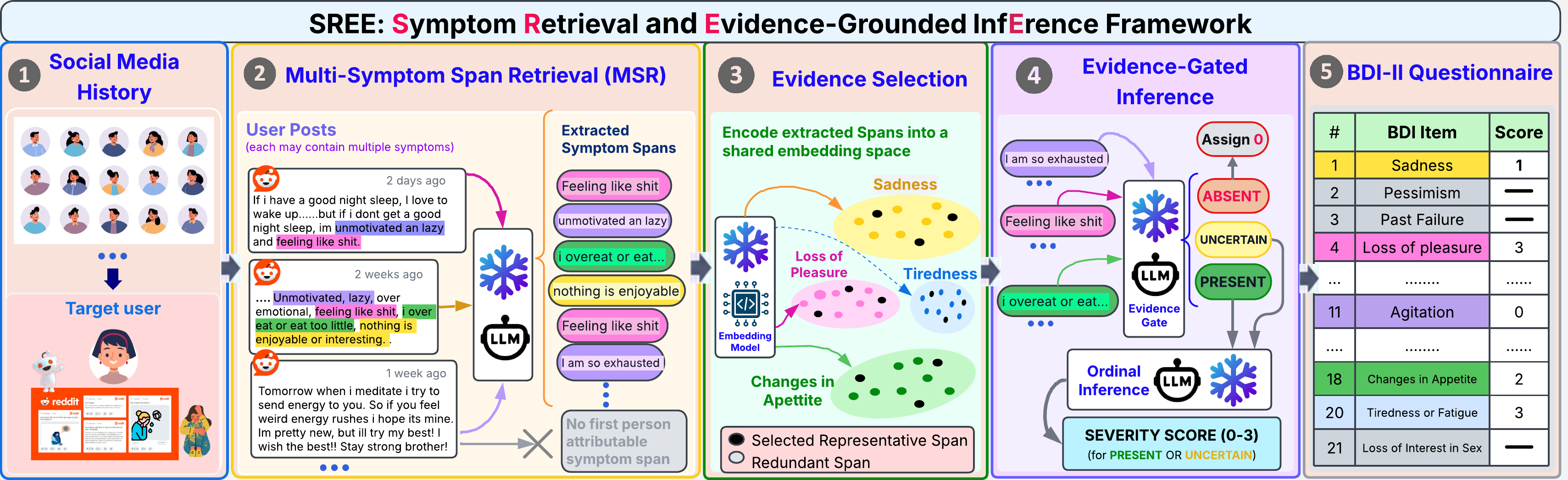}
    \caption{SREE framework for evidence-grounded BDI-II inference from longitudinal social-media histories.}
    \label{fig:sree_framework}
\end{figure*}

We  argue that questionnaire-guided assessment requires moving from \textit{semantic retrieval to attributable evidence localization.} Rather than treating an entire post as evidence, we use a large language model (LLM) to identify the first-person sub-span in the post that expresses a BDI-II symptom. Each extracted span is grounded to its source post, forming a user--item evidence pool. But identifying valid evidence is only the first step. Longitudinal histories may contain repeated instances of the same symptom, and exposing all of them to the scorer can distort severity inference. 
To address this issue, we use a Facility Location-inspired submodular objective \cite{lin-bilmes-2011-class} to select a compact subset of representative evidence. Evidence-Gated Inference then determines whether this subset provides sufficient attributable support for ordinal severity prediction and prevents residual, non-attributable evidence from affecting the predicted severity.

Social-media histories are not structured clinical interviews. Users may not discuss every BDI-II symptom, and symptom visibility varies across social-media language \cite{rissola2025incidence, tabusca-etal-2026-visibility, bao2026induction}. Prior methods typically require predictions for all 21 BDI-II items, implicitly treating the absence of retrieved evidence as an indication of symptom absence. This conflates missing evidence with evidence that a symptom is absent. To avoid this conflation, we adopt evidence-conditioned inference as our primary setting: a BDI-II item is scored only when supported by grounded linguistic evidence; otherwise, it remains unobserved. In clinical practice, unresolved items can be deferred for direct follow-up. For comparison with prior work, we report full-21-item reconstruction in Appendix~\ref{app:full21}.




Our Symptom Retrieval and Evidence-Grounded InfErence (SREE) framework\footnote{Code available at
\url{https://anonymous.4open.science/r/SREE_anonymous-C88E/}.} is illustrated in Figure~\ref{fig:sree_framework}. We evaluate attributable evidence retrieval on BDI-Sen at two levels of granularity: multi-symptom sentences and posts. We then evaluate evidence-conditioned BDI-II inference across eRisk 2019--2021. Our contributions are fourfold:




\noindent \textbullet\ We introduce Multi-Symptom Span Retrieval (MSR), which uses an LLM to extract multiple first-person symptom spans from a single post, moving beyond post-level semantic relevance.


\noindent \textbullet\ We adapt a Facility Location-inspired submodular objective to select a compact, representative subset of extracted evidence spans while reducing redundancy.


\noindent \textbullet\ We introduce Evidence-Gated Inference, which verifies the sufficiency of attributable symptom evidence before severity prediction, preventing unsupported residual content from influencing the inference process.


\noindent \textbullet\ We demonstrate that SREE achieves the highest exact item-level agreement (AHR) across all three eRisk collections.

\section{Related Work}

Computational mental-health research has increasingly moved from direct depression prediction to clinically grounded modeling, where symptoms and standardized psychological scales are used to structure inference and improve interpretability \cite{nguyen-etal-2022-improving, zhang2022psychiatricscaleguidedrisky}. This direction was further formalized in the CLEF eRisk shared tasks, which framed depression severity estimation as the reconstruction of responses to the 21 items of the Beck Depression Inventory-II (BDI-II) from users' social-media histories \cite{losada2019overview, perez2022automatic}. Together, these efforts shift the problem from assigning a single depression label toward symptom-level assessment.

A substantial body of work has reduced the gap between long user histories and item-level BDI-II inference by retrieving content relevant to each BDI-II item. Early approaches ranked posts using semantic similarity to BDI-II exemplars \cite{perez2023semantic}. This retrieve-then-infer paradigm was later extended through questionnaire-focused summarization, adaptive retrieval, and retrieval-augmented LLM reasoning, with downstream scoring ranging from trained predictors to LLMs \cite{aragon2024delving, wang2024explainable,  ravenda2025llms, ravenda2025transforming}. Recent work adopts multi-stage retrieval, using previously identified symptom evidence to guide subsequent evidence selection \cite{xu2026social}. Collectively, these studies establish retrieval as a central mechanism for transforming long social-media histories into symptom-specific evidence for questionnaire inference.

Parallel work has examined symptom evidence at a finer granularity through resources such as BDI-Sen and DepreSym, which link social-media sentences to individual BDI-II symptoms and support sentence-level evidence detection and explanation \cite{perez2023bdi, bao2024explainable, perez2025depresym}. Recent work further shows that symptom evidence is difficult to characterize consistently, that its presence or abundance does not necessarily translate into reliable severity estimation, and that linguistic observability varies substantially across BDI-II symptoms \cite{bao2026induction, rissola2025incidence, tabusca-etal-2026-visibility}. Prior work has largely focused on retrieving posts or sentences relevant to individual questionnaire items and determining whether they contain symptom evidence. We focus instead on the retrieval--scoring interface: identifying symptom-bearing sub-spans, constructing a grounded evidence set for each user--item pair, and examining how the composition and redundancy of that evidence affect BDI-II severity inference.

\section{Methodology}


\subsection{Problem Formulation}
\label{sec:problem_formulation}

Let $U$ denote a set of users. For each user $u_i\in U$, let $P_i=\{p_i^1,p_i^2,\ldots,p_i^{T_i}\}$ denote the user's longitudinal posting history, where $p_i^t$ is the $t$-th post authored by $u_i$. Let $B=\{b_1,b_2,\ldots,b_{21}\}$ denote the 21 BDI-II items, where each item $b_k$ consists of a symptom name, a description, and  four textual response choices $C_k=\{c_{k,0},c_{k,1},c_{k,2},c_{k,3}\}$, corresponding to increasing levels of symptom severity. For each user--item pair $(u_i,b_k)$, the target is the ordinal response $y_i^k\in\{0,1,2,3\}$ corresponding to one of these choices. The complete questionnaire response for user $u_i$ is therefore $\mathbf{y}_i=(y_i^1,y_i^2,\ldots,y_i^{21}).$ The total BDI-II score is $Y_i=\sum_{k=1}^{21} y_i^k$, which can be mapped to the eRisk categories: minimal ($0$--$9$), mild ($10$--$18$), moderate ($19$--$29$), and severe ($30$--$63$).

\subsection{SREE: Symptom Retrieval and Evidence-Grounded Inference}
Our central assumption is that reliable item-level inference should be
conditioned on explicit linguistic evidence from the user's own posting
history. Rather than treating an entire retrieved post as the evidence
unit, we represent evidence for item $b_k$ using grounded first-person
sub-spans anchored to their source posts. The resulting task has two coupled components. First, \emph{symptom
evidence retrieval} identifies which BDI-II items are expressed in a
user's posts and extracts the supporting grounded sub-spans. Second,
\emph{evidence-grounded inference} maps the retrieved evidence to an
item-level prediction $\hat{y}_i^k$.
Thus, the objective is
not merely to retrieve topically related posts, but to recover specific evidence
to support ordinal BDI-II item inference.

\subsubsection{Multi-symptom Span Retrieval (MSR)}
\label{sec:span_extraction}

For each post $p_i^t\in P_i$, a frozen zero-shot extractor
$f_{\mathrm{ext}}$ returns candidate tuples $(\hat{b},s,\kappa)$, where
$\hat{b}$ is the predicted BDI-II item, $s$ is the supporting textual
span, and $\kappa$ $\in [0,1]$ is the extraction confidence. The extractor identifies the shortest first-person span expressing the target BDI-II item, excluding others’ experiences, advice, quotations, and generic discussions. We discard malformed candidates with $\hat{b}\notin B$, missing supporting
spans, or confidence values outside $[0,1]$. To prevent
unsupported or hallucinated text from entering subsequent stages, each
remaining span must be grounded in its source post. After normalizing case, quotation marks, and whitespace, we retain a span only if its normalized text occurs as a contiguous substring ($\preceq$) of the
normalized source post. For each user
$u_i$ and BDI-II item $b_k$, we further require the extraction confidence
to meet or exceed a threshold $\tau$. Duplicate $(p_i^t,s)$ pairs are removed before downstream processing. The resulting post--span pairs form the grounded evidence pool \(V_{i,k}\):
\begin{equation*}
\begin{aligned}
V_{i,k}=\{(p_i^t,s)\mid\;&
(\hat b,s,\kappa)\in f_{\mathrm{ext}}(p_i^t),\;  \hat b=b_k,\\
&\mathrm{norm}(s)\preceq\mathrm{norm}(p_i^t),\;  \kappa\ge\tau\}.
\end{aligned}
\end{equation*}


\subsubsection{Evidence Selection}
\label{sec:facility_location}

The grounded evidence pool $V_{i,k}$ may contain multiple spans expressing similar symptom evidence, as users may repeatedly describe the same experience across posts. For example, for the BDI-II item \emph{Tiredness or Fatigue}, spans such as ``I feel exhausted every day,'' ``I am tired all the time,'' and ``I have no energy anymore'' may all provide valid evidence for the same underlying symptom. Retaining all such spans can over-represent repeated evidence while adding no new information for downstream scoring. We therefore seek a subset that preserves the distinct symptom evidence present in the pool, while avoiding unnecessary repetition.

For user--item pairs with $V_{i,k}\neq\emptyset$, we use a Facility
Location-inspired submodular objective \cite{lin-bilmes-2011-class} to
select a smaller representative subset that favors spans relevant to the
target BDI-II item while collectively representing the broader evidence
pool. Let $e=(p_i^t,s)\in V_{i,k}$ be an evidence unit, and let
$\phi(\cdot)$ denote an L2-normalized representation produced by a
pretrained sentence embedding model. Eq.~\eqref{eq:relevance} defines the
relevance score $R(s)$ as the mean of item relevance $R_I(s)$ and
response-choice relevance $R_C(s)$.
Specifically, $R_I(s)$ is the similarity between the textual representation of span $s$ and the item
$b_k$, while $R_C(s)$ is the maximum
similarity between $s$ and the four textual item response choices $c_{k,r}$, where
$r\in\{0,1,2,3\}$.
\begin{equation}
\label{eq:relevance}
\begin{aligned}
R(s) &= \frac{R_I(s)+R_C(s)}{2},
\quad [x]_+=\max(x,0)\\
R_I(s) &= \left[\phi(s)^\top\phi(b_k)\right]_+,\\
R_C(s) &= \max_{r\in\{0,1,2,3\}}
\left[\phi(s)^\top\phi(c_{k,r})\right]_+
\end{aligned}
\end{equation}

To reduce redundant evidence while preserving distinct symptom expressions, we seek a representative subset $S\subseteq V_{i,k}$ using the Facility Location-inspired objective in Eq.~\eqref{eq:facility}:
\begin{equation}
\label{eq:facility}
F(S)=
\sum_{e\in V_{i,k}}
\max_{e'\in S}
R(s_{e'})
\left[\phi(s_e)^\top\phi(s_{e'})\right]_+
\end{equation}
where $s_e$ denotes the span associated with evidence unit $e$, and $F(\emptyset)=0$. For each $e\in V_{i,k}$, the maximization selects the evidence unit $e'\in S$ whose span $s_{e'}$ best represents $e$ through relevance-weighted semantic similarity. Thus, $F(S)$ favors individually relevant spans that collectively represent the evidence in $V_{i,k}$. We construct $S$ greedily from $S_0=\emptyset$. At iteration $\ell$, each unselected unit $e\in V_{i,k}\setminus S_{\ell-1}$ forms a candidate extension $S_{\ell-1}\cup\{e\}$, and we select the unit  $e_\ell$ with the largest marginal gain $\Delta_F(e\mid S_{\ell-1})$, as defined in Eq.~\eqref{eq:marginal_gain}:

\begin{equation}
\label{eq:marginal_gain}
\begin{aligned}
\Delta_F(e\mid S)
&=F(S\cup\{e\})-F(S),\\
e_\ell
&=\arg\max_{e\in V_{i,k}\setminus S_{\ell-1}}
\Delta_F(e\mid S_{\ell-1}).
\end{aligned}
\end{equation}
Selection continues until the marginal gain $\Delta_F(e_\ell\mid S_{\ell-1})$ falls below a fraction \(\gamma\) of the initial gain $\Delta_F(e_1\mid S_0)$, subject to lower and upper bounds on the number of selected evidence units. The final evidence set
for user $u_i$ and item $b_k$ is denoted $S_{i,k}^{\ast}$.

\subsubsection{Evidence-Gated Inference}
\label{sec:evidence_gate_inference}

Given the representative evidence subset $S_{i,k}^{\ast}$, we next determine whether it is sufficient to support inference for the target BDI-II item. Social-media posts often mix self-reports with descriptions of other people, quoted content, hypothetical statements, or general discussion. Such cases can remain even though the extractor is explicitly prompted to identify first-person symptom evidence. For example, the extractor associated ``they get really violent'' with \emph{Agitation} in a post describing children at a mental-health clinic where the user works; ``I dont really have any idea what I'm talking about'' with \emph{Worthlessness} in a tax-advice context; and ``Punished for our sins'' with \emph{Punishment Feelings} in a general theological discussion.


To distinguish attributable symptom evidence from residual unsupported content, we apply an \emph{evidence gate} before ordinal inference. The gate receives the item $b_k$, its four response choices $C_k$, and the selected evidence subset
$S_{i,k}^{\ast}$, including each selected sub-span and its source post.
We denote this structured prompt by
$\mathcal{E}_{i,k}=(b_k,C_k,S_{i,k}^{\ast})$. The gate jointly evaluates $\mathcal{E}_{i,k}$ and outputs a gate
decision $g_{i,k}\in\{\textsc{Absent},\textsc{Uncertain},\textsc{Present}\}$. \textsc{Present} indicates that at least one selected evidence unit describes the user's own state for item $b_k$, while \textsc{Absent} indicates no such support. Ambiguous evidence is labeled \textsc{Uncertain}. Since evidence is evaluated jointly, references to others in one post do not imply \textsc{Absent} when another selected post provides clear first-person support. If $g_{i,k}=\textsc{Absent}$, the item response $\hat{y}_i^k$ is set to $0$ and no ordinal inference is performed. Otherwise, the same structured prompt $\mathcal{E}_{i,k}$ is passed to
the direct inference function $f_{\mathrm{dir}}$, which predicts a
response in $\{0,1,2,3\}$. For user--item pairs with $V_{i,k}\neq\emptyset$, Eq.~\eqref{eq:gated_inference} defines the resulting prediction:
\begin{equation}
\label{eq:gated_inference}
\hat{y}_i^k =
\begin{cases}
0, & g_{i,k}=\textsc{Absent},\\[1mm]
f_{\mathrm{dir}}\!\left(\mathcal{E}_{i,k}\right),
& \text{otherwise}.
\end{cases}
\end{equation}

If $V_{i,k}=\emptyset$, no grounded evidence is available for that user--item pair, and the item remains unobserved in the evidence-conditioned setting.

\begin{table}[t]
\centering
\footnotesize
\renewcommand{\arraystretch}{0.95}
\setlength{\tabcolsep}{3.5pt}

\begin{tabular}{@{}lcccc@{}}
\toprule
\textbf{Method}
& \shortstack{\textbf{Avg.}\\\textbf{Posts$\downarrow$}}
& \textbf{P$\uparrow$}
& \textbf{R$\uparrow$}
& \textbf{F1$\uparrow$} \\
\midrule

Fixed Top-$k$ Retrieval
& \multirow{2}{*}{147.45}
& \multirow{2}{*}{0.09}
& \multirow{2}{*}{\underline{0.69}}
& \multirow{2}{*}{0.16} \\
\cite{wang2024explainable}
& & & & \\[1pt]

Adaptive-$k^{\star}$ Retrieval
& \multirow{2}{*}{123.45}
& \multirow{2}{*}{0.10}
& \multirow{2}{*}{0.65}
& \multirow{2}{*}{0.16} \\
\cite{ravenda2025llms}
& & & & \\[1pt]

Two-Hop Retrieval
& \multirow{2}{*}{142.60}
& \multirow{2}{*}{0.10}
& \multirow{2}{*}{\textbf{0.73}}
& \multirow{2}{*}{0.17} \\
\cite{xu2026social}
& & & & \\

\midrule

\rowcolor{oursrow}
\textbf{MSR} \textcolor{blue}{(Gemma-3-4B-IT)}
& \underline{121.80}
& \underline{0.15}
& \textbf{0.73}
& \underline{0.24} \\ [3pt]

\rowcolor{oursrow}
\textbf{MSR} \textcolor{blue}{(Gemma-3-27B-IT)}
& \textbf{86.00}
& \textbf{0.17}
& 0.68
& \textbf{0.27} \\

\bottomrule
\end{tabular}

\caption{
Post-level retrieval on eRisk 2019 using BDI-Sen annotations.
P/R/F1 are macro-averaged across users. All reproduced baselines use
\texttt{all-MiniLM-L12-v2}. Shaded rows denote MSR (ours).
}
\label{tab:erisk_bdisen_retrieval_main}
\end{table}

\section{Experimental Design}
\label{sec:experimental_design}

\noindent \textbf{Datasets.} We evaluate SREE on the eRisk 2019 \cite{losada2019overview}, 2020 \cite{inbook2020}, and 2021 \cite{inbook2021} collections. These data provide
longitudinal Reddit histories with user-level BDI-II responses for
downstream questionnaire inference. Additionally, we evaluate
symptom-specific evidence retrieval on \textsc{BDI-Sen}
\cite{perez2023bdi}, derived from eRisk 2019 with sentence-level BDI-II
symptom annotations. Dataset statistics are provided in Appendix~\ref{app:dataset_stats}.

\noindent \textbf{Baselines.}
We compare against prior LLM-assisted questionnaire-guided methods that
retrieve, filter, or transform user evidence before BDI-II inference.
These include fixed top-$k$ retrieval \cite{wang2024explainable},
BDI-biased summarization \cite{aragon2024delving}, adaptive RAG (aRAG)
\cite{ravenda2025llms}, and AdaDepression's train-seeded two-hop
retrieval \cite{xu2026social}. These methods differ primarily in how
evidence is constructed and presented to the LLM for item-level
inference, aligning with our focus on evidence construction. We exclude supervised non-LLM classifiers as primary baselines because they rely on
task-specific training or engineered features. A tabulated comparison of methodologies is provided in Appendix~\ref{app:comp_methods}.

\noindent \textbf{Experimental Setup.}
We use \textsc{Gemma-3-27B-IT} \cite{gemma_2025} as the fixed language model throughout SREE, including MSR, evidence gating, and ordinal BDI-II inference. All LLM components operate zero-shot, without task- or dataset-specific fine-tuning, and use greedy decoding with
temperature $0$. We also reproduce all comparison baselines and evaluate
their retrieved evidence using the \textsc{Gemma-3-27B-IT} scorer.
Holding the language model fixed across methods reduces confounding from
differences in model capacity or choice, allowing the
comparison to focus on how evidence is retrieved, represented, and
attributed to individual BDI-II items. Prompt templates are provided in Appendix~\ref{app:prompts}.

\noindent \textbf{Parameter Selection.}
We use a single operating point for all downstream eRisk experiments.
MSR uses a confidence threshold of $\tau=0.70$. Facility Location uses
\texttt{all-MiniLM-L12-v2} \cite{reimers-2019-sentence-bert} span representations, and computes item/response-choice relevance  as in Eq.~\eqref{eq:relevance}.
Greedy selection stops when the marginal gain in
Eq.~\eqref{eq:marginal_gain} falls below $\gamma=0.20$ of the initial
marginal gain, with $k_{\min}=1$ and $k_{\max}=20$.
Evidence-Gated Inference uses the three-way
\textsc{Absent}-\textsc{Uncertain}-\textsc{Present} decision
defined in Section~\ref{sec:evidence_gate_inference}, with no additional
confidence threshold. Further details on parameter selection are
provided in Appendix~\ref{app:parameters}.

\noindent \textbf{Evaluation Metrics.}
Following the official eRisk evaluation framework and prior BDI-II
inference studies, we use Average Hit Rate (AHR), Average Closeness Rate
(ACR), and Average Difference between Overall Depression Levels (ADODL).
AHR measures exact item-level agreement, ACR captures ordinal distance
between predicted and gold responses, and ADODL compares aggregate BDI-II
scores. Following prior work \cite{ravenda2025llms,xu2026social}, we also
merge the directional alternatives for items 16 (\emph{Changes in Sleeping
Pattern}) and 18 (\emph{Changes in Appetite}) at each severity level
(e.g., 1a/1b, 2a/2b, and 3a/3b), preserving the $0$--$3$ ordinal scale. Our primary downstream evaluation is \emph{evidence-conditioned}: we score only user--item pairs with grounded symptom evidence and compare all baselines on the same supported pairs. Accordingly, ADODL is adapted to the number of evaluated items, using a maximum score of $3m_i$ rather than $63$, where $m_i$ denotes the number of evaluated items for user $u_i$. We omit Depression Category Hit Rate
(DCHR), since it requires a complete 21-item
questionnaire. Full metric definitions and evidence-conditioned
adaptations are provided in Appendix~\ref{app:full21}.

\begin{table}[t]
\centering
\footnotesize
\setlength{\tabcolsep}{3.8pt}
\renewcommand{\arraystretch}{0.95}

\begin{tabular}{@{}lcccc@{}}
\toprule
\textbf{Method}
& \textbf{Cov.}
& \textbf{P$\uparrow$}
& \textbf{R$\uparrow$}
& \textbf{F1$\uparrow$} \\
\midrule

all-mpnet-base-v2$^{\textcolor{blue}{a, c}}$
& 0.42
& 0.56
& 0.38
& 0.45 \\

all-MiniLM-L12-v2$^{\textcolor{blue}{b}}$
& 0.42
& 0.56
& 0.38
& 0.45 \\

msmarco-MiniLM-L12-v3$^{\textcolor{blue}{c}}$
& 0.39
& 0.54
& 0.37
& 0.44 \\

msmarco-distilbert-base-v4$^{\textcolor{blue}{c}}$
& 0.40
& 0.55
& 0.37
& 0.44 \\

\midrule

\rowcolor{oursrow}
\textbf{MSR (Gemma-3-27B-IT)}
& 0.39
& \textbf{0.68}
& \textbf{0.40}
& \textbf{0.50} \\

\bottomrule
\end{tabular}

\vspace{2pt}

\parbox{\columnwidth}{
\scriptsize
$^{a}$~APB~\cite{bascunana2024apb}; 
$^{b}$~Two-Hop~\cite{xu2026social};
$^{c}$~Adaptive RAG~\cite{ravenda2025llms}.
}

\caption{
Multi-symptom BDI-II item retrieval on 144 BDI-Sen examples
under the matched-budget setting ($K=2$). Best results are boldfaced.
}
\label{tab:multisymptom}
\end{table}

\noindent \textbf{Research Questions (RQs).}
We evaluate SREE along two dimensions. \textbf{RQ1:} Does MSR provide higher-quality and attributable BDI-II evidence than existing retrieval and evidence-construction strategies? \textbf{RQ2:} Can SREE support accurate item-level BDI-II inference under the evidence-conditioned setting across the eRisk collections?










\begin{table*}[t]
\centering
\footnotesize
\renewcommand{\arraystretch}{0.9}

\begin{tabularx}{0.97\textwidth}{
    @{}
    >{\raggedright\arraybackslash}p{1.9cm}
    >{\raggedright\arraybackslash}X
    >{\centering\arraybackslash}p{1.35cm}
    >{\centering\arraybackslash}p{1.35cm}
    >{\centering\arraybackslash}p{2.90cm}
    @{}
}
\toprule

\multirow{2}{*}{\textbf{Collection}}
&
\multirow{2}{*}{\textbf{Methods}}
&
\multicolumn{2}{c}{\textbf{Symptom Metrics}}
&
\multicolumn{1}{c}{\textbf{Questionnaire Metric}}
\\

\cmidrule(lr){3-4}
\cmidrule(l){5-5}

&
&
\textbf{AHR$\uparrow$}
&
\textbf{ACR$\uparrow$}
&
\textbf{ADODL$\uparrow$}
\\

\midrule


\multirow{3}{*}{
\makecell[l]{\textbf{eRisk 2019}\\
20 users \\ 327 items}
}
&
BDI-biased Summarization~\cite{aragon2024delving}
&
30.55
&
59.31
&
72.00
\\

&
Fixed Top-$k$ Retrieval~\cite{wang2024explainable}
&
25.99
&
68.93
&
77.40
\\

&
Adaptive RAG (aRAG)~\cite{ravenda2025llms}
&
\underline{33.14}
&
\textbf{71.30}
&
\textbf{83.18}
\\

\noalign{\vskip 1.5pt}
\cdashline{2-5}
\noalign{\vskip 3.5pt}

\rowcolor{oursrow}
\textbf{Ours}
&
\textbf{SREE Framework}
&
\makecell{\textbf{36.13}\\\posdelta{2.99}}
&
\makecell{\underline{69.21}\\\negdelta{2.09}}
&
\makecell{\underline{78.37}\\\negdelta{4.81}}
\\

\midrule


\multirow{3}{*}{
\makecell[l]{\textbf{eRisk 2020}\\
70 users \\ 946 items}
}
&
BDI-biased Summarization~\cite{aragon2024delving}
&
30.39
&
61.01
&
72.32
\\

&
Fixed Top-$k$ Retrieval~\cite{wang2024explainable}
&
32.15
&
\textbf{71.73}
&
\underline{80.31}
\\

&
Adaptive RAG (aRAG)~\cite{ravenda2025llms}
&
\underline{33.84}
&
\underline{71.48}
&
80.15
\\

\noalign{\vskip 1.5pt}
\cdashline{2-5}
\noalign{\vskip 3.5pt}

\rowcolor{oursrow}
\textbf{Ours}
&
\textbf{SREE Framework}
&
\makecell{\textbf{34.22}\\\posdelta{0.38}}
&
\makecell{70.30\\\negdelta{1.43}}
&
\makecell{\textbf{81.19}\\\posdelta{0.88}}
\\

\midrule


\multirow{4}{*}{
\makecell[l]{\textbf{eRisk 2021}\\
79 users \\ 1064 items}
}
&
BDI-biased Summarization~\cite{aragon2024delving}
&
23.19
&
55.92
&
66.93
\\

&
Fixed Top-$k$ Retrieval~\cite{wang2024explainable}
&
\underline{34.29}
&
\textbf{72.68}
&
\textbf{83.30}
\\

&
Adaptive RAG (aRAG)~\cite{ravenda2025llms}
&
33.01
&
70.36
&
79.93
\\

&
AdaDepression~\cite{xu2026social}
&
29.37
&
68.34
&
77.05
\\
\noalign{\vskip 1.5pt}
\cdashline{2-5}
\noalign{\vskip 3.5pt}

\rowcolor{oursrow}
\textbf{Ours}
&
\textbf{SREE Framework}
&
\makecell{\textbf{36.30}\\\posdelta{2.01}}
&
\makecell{\underline{70.79}\\\negdelta{1.89}}
&
\makecell{\underline{83.28}\\\negdelta{0.02}}
\\

\bottomrule
\end{tabularx}

\caption{
Evidence-conditioned BDI-II inference results on the eRisk collections (in \%).
\textbf{Bold} indicates the best result and \underline{underline} indicates
the second-best result within each collection.
For \textbf{Ours}, the value shown below each score denotes
\textbf{$\Delta=\text{Ours}-\text{best competing baseline}$} for that metric.
Higher values are better.
}
\label{tab:evidence_only_results}

\end{table*}

\section{Experimental Results}

\subsection{RQ1: Symptom Evidence Retrieval}
\label{sec:rq1_retrieval}

We evaluate evidence retrieval at two levels: post-level recovery of BDI-Sen-positive posts from full eRisk histories, and multi-symptom sentence-level recovery of labeled BDI-II items. 


\noindent \textbf{Post-level evidence retrieval.} We extend the post-level retrieval analysis of \citet{xu2026social} by reproducing prior retrieval strategies and adding MSR under the same evaluation setting in Table~\ref{tab:erisk_bdisen_retrieval_main}. At the macro level, Two-Hop achieves higher recall (0.73), whereas MSR
achieves higher precision (0.17) and F1 (0.27) while retaining
substantially fewer posts (86 vs.\ 142.60 per user). aRAG adaptively selects how many posts to retrieve, yet still retains 123.45 posts per user with lower precision and F1 than MSR. Moreover, nearly 75\% of its false-positive posts are excluded by
MSR, while MSR recovers about 29\% more gold-positive posts. This difference reflects how evidence is licensed: aRAG selects posts by relative semantic proximity to BDI-II response text, whereas MSR requires a grounded first-person symptom span before a post
contributes evidence. As a result, posts with overlapping symptom-related vocabulary, such as benign references to the future, treatment advice, or medication
discussion, can rank highly despite not expressing the target symptom
as a first-person self-report. The gain therefore lies in evidence selectivity and attribution rather than a smaller retrieval budget. Additional diagnostics and qualitative examples are provided in Appendix~\ref{app:retrieval_analysis}.

\noindent \textbf{Multi-symptom evidence retrieval.}
We compare MSR with four dense sentence encoders representing the semantic-retrieval families used in prior BDI-II pipelines. For each multi-symptom BDI-Sen sentence, dense methods rank BDI-II items by cosine similarity, while MSR produces its natural grounded span prediction set. Table~\ref{tab:multisymptom} reports results on 144 multi-symptom examples under the matched budget ($K=2$). Coverage@$K$ uses the top $\min(K,|\mathrm{pred}|)$ predictions; dense P/R/F1 are computed at $K$, whereas MSR is evaluated on its unpadded prediction set. Under $K=2$, MSR achieves the highest precision, recall, and F1 (0.68/0.40/0.50), compared with a best dense F1 of 0.45, while emitting only 1.74 predictions per example. This advantage persists when the dense retrieval budget is doubled: additional candidates increase dense recall but also introduce more unsupported labels, while MSR maintains a smaller set of predictions grounded in attributable textual spans. Results under the $2\times$ dense budget ($K=4$) are reported in Appendix~\ref{app:retrieval_analysis}.

\begin{table*}[t]
\centering
\footnotesize
\renewcommand{\arraystretch}{0.92}
\setlength{\tabcolsep}{7pt}

\begin{tabular}{
    @{}
    l
    @{\hspace{11pt}}
    l
    c
    c
    c
    @{}
}

\toprule

\textbf{Collection}
&
\textbf{Ablation Setting}
&
\textbf{AHR$\uparrow$}
&
\textbf{ACR$\uparrow$}
&
\textbf{ADODL$\uparrow$}
\\

\midrule


\multirow{4}{*}{
\makecell[l]{
\textbf{eRisk 2019}\\
20 users \\ 327 items
}}


&
\makecell[l]{
Keep-all + Ordinal Inference
(\textcolor{blue}{\textit{w/o Facility Location \& Evidence Gate}})
}
&
27.28
&
65.87
&
73.78
\\

\noalign{\vskip 1.5pt}

&
\makecell[l]{
Keep-all + Evidence-Gated Inference
(\textcolor{blue}{\textit{w/o Facility Location}})
}
&
28.85
&
66.12
&
74.22
\\

\noalign{\vskip 1.5pt}

&
\makecell[l]{
Facility Location + Ordinal Inference
(\textcolor{blue}{\textit{w/o Evidence Gate}})
}
&
\underline{34.48}
&
\textbf{69.22}
&
\underline{77.82}
\\

\noalign{\vskip 1.5pt}

&
\cellcolor{oursrow}
\textbf{Facility Location + Evidence-Gated Inference}
&
\cellcolor{oursrow}\textbf{36.13}
&
\cellcolor{oursrow}\underline{69.21}
&
\cellcolor{oursrow}\textbf{78.37}
\\

\midrule


\multirow{4}{*}{
\makecell[l]{
\textbf{eRisk 2020}\\
70 users \\ 946 items
}}

&
\makecell[l]{
Keep-all + Ordinal Inference
(\textcolor{blue}{\textit{w/o Facility Location \& Evidence Gate}})
}
&
31.39
&
68.73
&
75.45
\\

\noalign{\vskip 1.5pt}

&
\makecell[l]{
Keep-all + Evidence-Gated Inference
(\textcolor{blue}{\textit{w/o Facility Location}})
}
&
31.09
&
68.54
&
76.26
\\

\noalign{\vskip 1.5pt}

&
\makecell[l]{
Facility Location + Ordinal Inference
(\textcolor{blue}{\textit{w/o Evidence Gate}})
}
&
\textbf{34.90}
&
\textbf{70.92}
&
\underline{80.11}
\\

\noalign{\vskip 1.5pt}

&
\cellcolor{oursrow}
\textbf{Facility Location + Evidence-Gated Inference}
&
\cellcolor{oursrow}\underline{34.22}
&
\cellcolor{oursrow}\underline{70.30}
&
\cellcolor{oursrow}\textbf{81.19}
\\

\midrule


\multirow{4}{*}{
\makecell[l]{
\textbf{eRisk 2021}\\
79 users \\ 1064 items
}}


&
\makecell[l]{
Keep-all + Ordinal Inference
(\textcolor{blue}{\textit{w/o Facility Location \& Evidence Gate}})
}
&
33.38
&
70.44
&
81.86
\\

\noalign{\vskip 1.5pt}

&
\makecell[l]{
Keep-all + Evidence-Gated Inference
(\textcolor{blue}{\textit{w/o Facility Location}})
}
&
34.23
&
70.29
&
82.39
\\

\noalign{\vskip 1.5pt}

&
\makecell[l]{
Facility Location + Ordinal Inference
(\textcolor{blue}{\textit{w/o Evidence Gate}})
}
&
\underline{35.26}
&
\textbf{71.23}
&
\underline{83.17}
\\

\noalign{\vskip 1.5pt}

&
\cellcolor{oursrow}
\textbf{Facility Location + Evidence-Gated Inference}
&
\cellcolor{oursrow}\textbf{36.30}
&
\cellcolor{oursrow}\underline{70.79}
&
\cellcolor{oursrow}\textbf{83.28}
\\

\bottomrule
\end{tabular}

\caption{Evidence-conditioned ablation of evidence composition and inference components on the eRisk collections. Shaded cells indicate the final SREE configuration used in the primary evaluation.}
\label{tab:evidence_only_ablation}
\end{table*}

\subsection{RQ2: Evidence-Conditioned BDI-II Inference}

We examine whether the grounded evidence constructed in RQ1, after Facility Location selection and Evidence-Gated Inference, supports accurate BDI-II response prediction. Table~\ref{tab:evidence_only_results} compares all methods on the same evidence-supported user--item pairs within each collection: 327 for eRisk 2019, 946 for 2020, and 1,064 for 2021. SREE achieves the highest AHR on all three collections: 36.13, 34.22, and 36.30, corresponding to gains of $+2.99$, $+0.38$, and $+2.01$  percentage points over the strongest competing baseline. Because AHR measures exact BDI-II response agreement, these results suggest that selecting representative evidence and restricting inference to sufficiently supported items can improve item-level response selection across collections, with varying gains.

The ordinal and questionnaire-level metrics are mixed. SREE ranks second on ACR in 2019 and 2021, trailing the best baseline by 2.09 and 1.89 points, and third in 2020 ($-1.43$). For ADODL, it ranks second in 2019, first in 2020 ($+0.88$), and is effectively tied for best in 2021 (83.28 vs.\ 83.30). Overall, SREE ranks first or second in eight of the nine collection--metric comparisons, with its clearest advantage in exact item-level agreement. Full-questionnaire evaluation, including unsupported items, is provided
in Appendix~\ref{app:full21}. We next analyze how representative evidence
selection and evidence sufficiency affect the observed gains and tradeoffs.

\begin{figure*}[t]
    \centering
    \includegraphics[width=\textwidth]{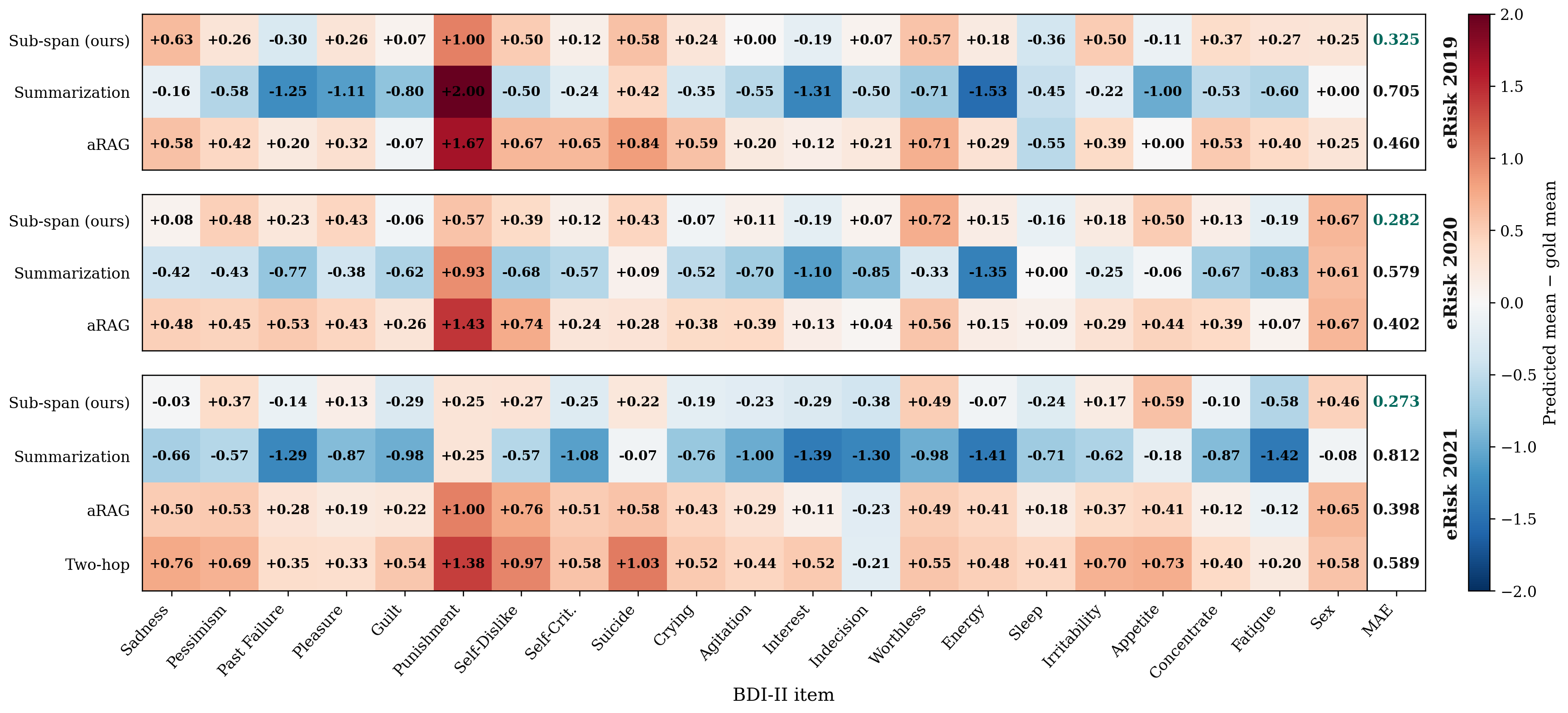}
    \caption{Population-level symptom calibration across eRisk 2019--2021. Each cell reports the difference between mean predicted and gold BDI-II scores over evidence-supported user--item pairs; positive values indicate overprediction and negative values underprediction. The rightmost column reports MAE across all 21 symptoms.}
    \label{fig:heatmap}
\end{figure*}

\section{Discussion}

\noindent \textbf{Evidence composition and sufficiency.}
The results above indicate that the benefits of SREE extend beyond retrieval. Downstream inference depends on how grounded evidence is composed and whether it sufficiently supports ordinal prediction. Table~\ref{tab:evidence_only_ablation} isolates these decisions while holding extraction, grounding, and ordinal inference fixed. \emph{Keep-all} passes all grounded evidence directly to inference, whereas \emph{Facility Location} forms a compact representative subset. \emph{Evidence-Gated Inference} then determines whether the selected evidence is sufficient, mapping \textsc{Absent} cases to 0 and otherwise applying the same ordinal inference.

Replacing Keep-all with Facility Location improves AHR by \(7.20\), \(3.51\), and \(1.88\) points on eRisk 2019--2021, respectively, while also improving ACR and ADODL in all three collections. Across years, the number of extracted spans is nearly uncorrelated with gold severity (\(\rho\approx 0.02\)--\(0.03\)), but substantially associated with Keep-all predictions (\(\rho\approx 0.48\)--\(0.51\)). Ordinal inference therefore appears sensitive to evidence multiplicity even when that multiplicity carries little information about the true response. Facility Location mitigates this by replacing the full evidence pool with a smaller representative set of distinct spans.


Evidence-Gated Inference addresses evidence sufficiency after selection. Its effect is smaller and less uniform: relative to Facility-only inference, it improves AHR in 2019 and 2021 but slightly reduces it in 2020, with only modest changes in ACR and ADODL. We therefore interpret the gate as a control on unsupported non-zero responses rather than a
uniform source of higher accuracy. Together, these results show that BDI-II inference depends not only on identifying relevant evidence, but also on how that evidence is composed and assessed before ordinal prediction.

\noindent \textbf{Population-level symptom calibration.} We next examine whether differences in evidence representation are reflected in the resulting population-level symptom profiles. As shown in Figure~\ref{fig:heatmap}, the green-highlighted grounded sub-span evidence achieves the lowest profile-level MAE across all three collections (\(0.325/0.282/0.273\)), compared against \(0.705/0.579/0.812\) for BDI-biased summarization and \(0.460/0.402/0.398\) for aRAG; Two-Hop reaches \(0.589\) on 2021. The signed errors further reveal distinct systematic tendencies across evidence representations. Summarization predominantly shifts item means downward, whereas aRAG and Two-Hop more often shift them upward. Grounded sub-span evidence is more balanced around the gold profile, although
item-specific errors remain, including overestimation of \emph{Punishment Feelings} in 2019 and Appetite in 2021, and underestimation of \emph{Fatigue} in
2021. Grounding therefore reduces broad directional bias rather than eliminating symptom-level error uniformly.

This perspective helps explain why AHR gains do not translate uniformly to ACR and ADODL. ACR rewards ordinal proximity, whereas ADODL aggregates item-level errors that may partially offset. Thus, methods can achieve similar ordinal scores while exhibiting different symptom-level biases. For example, on eRisk 2021, grounded sub-span evidence and aRAG obtain similar ACR (\(70.79\) vs.\ \(70.36\)), yet aRAG shows predominantly positive bias while the grounded profile remains closer to zero. These patterns suggest that grounded evidence better preserves the observed symptom-severity profile beyond exact matching.

\section{Conclusion}


We studied evidence-grounded BDI-II inference from longitudinal
social-media histories, focusing on the gap between semantically relevant content and attributable symptom evidence. SREE uses frozen zero-shot LLM components to localize first-person symptom-specific sub-spans, selecting
a compact representative set with Facility Location, and applying
Evidence-Gated Inference before ordinal prediction. Across BDI-Sen and eRisk collections, MSR provided more
selective symptom evidence \textbf{(RQ1)}, while the complete pipeline
achieved the highest exact item-level agreement across all three eRisk
collections and remained competitive on ordinal and aggregate metrics
\textbf{(RQ2)}. Further analysis showed that representative selection
reduces the influence of redundant evidence, while grounding better
preserves population-level symptom profiles. Overall, these findings
suggest that questionnaire inference from longitudinal text is not only
a retrieval problem, but also an evidence-construction problem shaped by
attribution, composition, and evidential support before inference.

\section{Limitations}

Our primary downstream evaluation is evidence-conditioned, leaving
BDI-II items without grounded evidence unobserved. This isolates
inference quality when textual evidence is available, but does not
measure complete questionnaire reconstruction. As shown in
Appendix~\ref{app:full21}, assigning unobserved items a score of 0 can
substantially affect questionnaire-level performance. 

Our evaluation is limited to the eRisk Reddit collections and BDI-Sen. These datasets contain relatively small user cohorts, and BDI-II symptoms differ in how visibly they are expressed in social-media language. The results may therefore not generalize
directly to other platforms, languages, or clinical populations. 

Finally, the reported SREE pipeline uses Gemma-3-27B-IT for MSR and
downstream questionnaire inference. Holding the model fixed across
reproduced baselines reduces confounding, but the results may still
depend on this model. MSR can also produce residual attribution errors
for semantically related but non-attributable text. 

The zero-shot extraction setting may limit the ability to capture implicit symptom evidence while maintaining reliable abstention. Future work could preserve that abstention while improving attribution, including adapting Symptom Induction~\cite{bao2026induction} from
sentence-level BDI-II relevance to span localization.

\section{Ethical Considerations}
This work is intended for research on evidence-grounded questionnaire
inference and not as a clinical diagnostic system. The ground-truth
BDI-II responses in the eRisk collections are questionnaires completed
by the social-media users themselves \cite{losada2019overview,inbook2020,inbook2021} and do not constitute clinical diagnoses. Accordingly, predictions produced by SREE should
not substitute for assessment by qualified mental-health professionals
or support high-stakes decisions such as suicide-risk intervention.
Our experiments use only the released and anonymized eRisk collections
and do not involve additional Reddit scraping, attempts to identify
users, or contact with individuals. We follow the access and usage
conditions of the datasets. Because social-media posts may contain sensitive personal disclosures, examples included in the paper are de-identified and paraphrased when
necessary to illustrate the methodology. Any future deployment based on this approach would require additional safeguards for privacy, clinical
validation, and responsible human oversight.


\bibliography{custom}
\appendix

\section{Comparison with Prior Questionnaire-Guided Approaches}
\label{app:comp_methods}

Table~\ref{tab:method_comparison} summarizes the main methodological
differences between SREE and prior questionnaire-guided approaches.

\begin{table*}[t]
\centering
\footnotesize
\setlength{\tabcolsep}{3.2pt}
\renewcommand{\arraystretch}{1.12}
\renewcommand{\tabularxcolumn}[1]{m{#1}}

\begin{tabularx}{0.97\textwidth}{
    @{}
    >{\raggedright\arraybackslash}m{3.5cm}
    >{\raggedright\arraybackslash}m{2.7cm}
    >{\centering\arraybackslash}m{1.2cm}
    >{\centering\arraybackslash}m{1.2cm}
    >{\centering\arraybackslash}m{1.5cm}
    >{\raggedright\arraybackslash}X
    >{\raggedright\arraybackslash}m{1.6cm}
    @{}
}
\toprule

\textbf{Method}
&
\makecell[l]{\textbf{Evidence Unit}}
&
\makecell{\textbf{Keyword}\\\textbf{Pre-filter}}
&
\makecell{\textbf{Dense}\\\textbf{Retrieval}}
&
\makecell{\textbf{LLM}\\\textbf{Extraction}}
&
\makecell[l]{\textbf{Evidence}\\\textbf{Selection}}
&
\makecell[l]{\textbf{Task}\\\textbf{Adaptation}}
\\

\midrule

Fixed Top-K Retrieval \newline \cite{wang2024explainable}
&
Top-\(k\) writings (Post)
&
\cmark
&
\cmark
&
\xmark
&
cos$\theta$ + top-$k$
&
--
\\

\cdashline{1-7}
\noalign{\vskip 2pt}

BDI-biased Summarization \newline \cite{aragon2024delving}
&
Generated BDI-II biased summary
&
\xmark
&
\cmark
&
\xmark
&
cos$\theta$ threshold \newline + BART
&
SAMSum-ft BART
\\

\cdashline{1-7}
\noalign{\vskip 2pt}

Adaptive RAG (aRAG) \newline \cite{ravenda2025llms}
&
Post
&
\xmark
&
\cmark
&
\xmark
&
\makecell[l]{ABIDE \\ Adaptive $k^{*}$}
&
--
\\

\cdashline{1-7}
\noalign{\vskip 2pt}

AdaDepression \newline \cite{xu2026social}
&
Post
&
\xmark
&
\cmark
&
\xmark
&
\makecell[l]{Train-seeded \\ two-hop}
&
Train users + RL
\\

\cdashline{1-7}
\noalign{\vskip 2pt}

\rowcolor{oursrow}
\textbf{SREE Framework} \newline (Ours)\textsuperscript{\dag}
&
\textbf{Symptom-specific Span}
&
\xmark
&
\xmark
&
\cmark
&
\textbf{Facility Location-inspired}
&
\textbf{--}
\\

\bottomrule
\end{tabularx}

\vspace{1pt}
\begin{minipage}{0.99\textwidth}
\scriptsize
\textsuperscript{\dag}
Prior methods infer all 21 BDI-II items and assign 0 when no evidence is retrieved.
Our evidence-conditioned setting instead abstains when no grounded sub-span is available.
Full-21 results are reported separately for comparison.
\textbf{--} means zero-shot for task adaptation.
\end{minipage}

\caption{
Methodological Comparison of LLM-Assisted Questionnaire-Guided Depression Screening Approaches
}
\label{tab:method_comparison}
\end{table*}

\section{Supplementary Experimental Details}
\label{app:supp_exp}

\paragraph{Dataset Statistics}
\label{app:dataset_stats}

Table~\ref{tab:datasets} summarizes the three eRisk collections used
for downstream BDI-II inference. The mean BDI-II total scores are
24.3, 22.5, and 28.4 for eRisk 2019 \cite{losada2019overview},
2020 \cite{inbook2020}, and 2021 \cite{inbook2021}, respectively.
The corresponding distributions across the minimal, mild, moderate,
and severe categories are reported in the table. For eRisk 2021, the
official collection contains 80 users. However, complete 21-item
BDI-II ground truth was unavailable for one user in the reproduced
evaluation files. We therefore exclude this user from all
questionnaire-level evaluations, resulting in 79 users. The same
79-user subset is used for the reproduced aRAG and AdaDepression
baselines, ensuring that all questionnaire comparisons use the same
user set. Table~\ref{tab:bdisen_stats} summarizes the annotation structure of
BDI-Sen \cite{perez2023bdi}. Table~\ref{tab:bdisen_examples} provides representative
examples of multi-positive, single-positive, and no-positive
annotations, illustrating that a sentence may be associated with
multiple symptom labels and that annotated symptoms may be either
positive (\textcolor{green!50!black}{\textbf{1}}) or negative (\textcolor{red!70!black}{\textbf{0}}). The eRisk collections (2019--2021) and BDI-Sen were obtained for research use through
their respective official dataset-access request procedures.

\paragraph{Relationship Between BDI-Sen and eRisk 2019}
BDI-Sen is derived from the eRisk 2019 collection and provides
sentence-level BDI-II symptom annotations, whereas eRisk 2019 provides
user-level posting histories together with BDI-II questionnaire
responses. The BDI-Sen files used in our experiments do not include
identifiers linking each sentence to an eRisk user or post. We construct
the post-level relevance set by normalized substring matching: after
lowercasing and whitespace normalization, a BDI-Sen sentence is aligned
to every 2019 post whose TITLE+TEXT contains that sentence as a
contiguous substring. Posts that contain at least one matched \emph{positive} (\texttt{label}$=1$) sentence form the symptom-bearing target set for post-level retrieval (441 unique posts). The labeled unit remains the sentence; the post is a retrieval target only because it contains that sentence. This post-level relevance set is used in the post-level retrieval evaluation in Section~\ref{sec:rq1_retrieval}, reported in
Table~\ref{tab:erisk_bdisen_retrieval_main}.

\begin{table}[t]
\centering
\small
\begin{tabular}{lrrr}
\toprule
Release Year & 2019 & 2020 & 2021 \\
\midrule
Users & 20 & 70 & 80 \\
Posts & 10,418 & 33,304 & 31,125 \\
Avg.\ posts / user & 520.9 & 475.8 & 389.1 \\
Avg.\ words / post & 46.4 & 42.6 & 49.3 \\
Minimal (0--9) & 4 & 10 & 6 \\
Mild (10--18) & 4 & 23 & 13 \\
Moderate (19--29) & 4 & 18 & 27 \\
Severe (30--63) & 8 & 19 & 34 \\
\bottomrule
\end{tabular}
\caption{Dataset statistics for the eRisk 2019, 2020, and 2021 BDI-II severity-assessment collections.}
\label{tab:datasets}
\end{table}

\noindent \textbf{Multi-Symptom Evaluation on BDI-Sen.}
Table~\ref{tab:bdisen_stats} shows that 550 BDI-Sen sentences contain at least one positive symptom label, and 151 contain more than one positive symptom label. After mapping to the 21 BDI-II symptom items, these counts become 545 and
144, respectively, because \emph{Social withdrawal} is not treated as
a separate BDI-II item in our mapping. We use the 144 sentences that
retain at least two distinct BDI-II items for the multi-symptom
retrieval evaluation reported in Table~\ref{tab:multisymptom}.

For each sentence $i$, let $G_i$ be the set of BDI-II items with a
positive BDI-Sen label after mapping onto the 21-item inventory, and let
$P_i$ be the items predicted for that sentence. BDI-Sen does not
annotate every item for every sentence. For this evaluation, all mapped
BDI-II items absent from $G_i$ are treated as negative decisions. Because BDI-Sen annotations were constructed from symptom-specific
candidate pools filtered by SBERT similarity, this yields a conservative
closed-world evaluation that penalizes predictions outside the original
annotation pool. Thus,
$TP_i=|G_i\cap P_i|$,
$FP_i=|P_i\setminus G_i|$, and
$FN_i=|G_i\setminus P_i|$.
Precision, recall, and F1 are computed from the pooled $TP$, $FP$, and
$FN$ counts over all 144 sentences and 21 items (micro-averaging).
For dense encoders, $P_i$ contains the top-$K$ items ranked by cosine
similarity to the item query, whereas for MSR, $P_i$ contains the
naturally emitted BDI-II item predictions ranked by extraction
confidence, without padding to $K$. We additionally report Coverage@$K$, defined as
\[
\mathrm{Coverage@}K =
\frac{1}{N}\sum_{i=1}^{N}
\frac{|G_i \cap P_i^{(K)}|}{|G_i|},
\]
where $N=144$ and $P_i^{(K)}$ contains the top
$\min(K,|P_i|)$ predicted items. Unlike micro-averaged recall,
Coverage@$K$ first computes the fraction of recovered gold items for
each sentence and then averages across sentences, giving each sentence
equal weight.

To provide a broader view of the zero-shot candidate-extraction stage
of MSR on BDI-Sen, Table~\ref{tab:bdisen_setting_results} reports
performance over three evaluation settings: the full 2,529-sentence
corpus, the 545 sentences containing at least one mapped positive
BDI-II item, and the 144 sentences containing at least two mapped
positive BDI-II items. Because this evaluation is zero-shot, all 2,529 BDI-Sen sentences are
used for evaluation without a dataset-specific train/validation/test
partition. These
results use the zero-shot candidate-extraction prompt in
Figure~\ref{fig:prompt}, with the item-name-plus-description symptom
inventory, at $\tau=0.70$. Precision, recall, and F1 are
micro-averaged over the sentence--item decisions. Precision increases substantially as the evaluation is restricted to sentences containing positive symptom
evidence, while recall changes little.

\begin{table}[t]
\centering
\small
\setlength{\tabcolsep}{6pt}
\renewcommand{\arraystretch}{1.05}

\begin{tabular}{lr}
\toprule
\textbf{Statistic} & \textbf{BDI-Sen} \\
\midrule
Labeled pairs & 5,003 \\
Unique sentences & 2,529 \\
$>$1 labeled symptom & 948 \\
$\geq$1 positive symptom & 550 \\
$>$1 positive symptom & 151 \\
\bottomrule
\end{tabular}

\caption{
Annotation statistics for BDI-Sen.
Labeled pairs denote sentence--symptom annotation entries.
}
\label{tab:bdisen_stats}
\end{table}

\begin{table*}[t]
\centering
\small
\setlength{\tabcolsep}{6pt}
\renewcommand{\arraystretch}{1.4}

\begin{tabularx}{0.97\textwidth}{
    @{}
    >{\raggedright\arraybackslash}p{2.1cm}
    >{\raggedright\arraybackslash}X
    >{\raggedright\arraybackslash}p{5.4cm}
    @{}
}
\toprule
\textbf{Type} &
\textbf{BDI-Sen sentence} &
\textbf{Symptom labels} \\
\midrule

Multi-positive
&
\textit{Yeah, mainly because Im tired all the time, no matter how much sleep I get.}
&
Changes in sleeping pattern
\textcolor{green!50!black}{\textbf{1}}; \newline
Tiredness or fatigue
\textcolor{green!50!black}{\textbf{1}}
\\

Single-positive
&
\textit{For the past few days, I have been completely exhausted.}
&
Tiredness or fatigue
\textcolor{green!50!black}{\textbf{1}}
\\

No positive
&
\textit{You're not doing anything wrong, it's not some personal defect,
and you're not at fault for going through this.}
&
Sadness \textcolor{red!70!black}{\textbf{0}};
Past failure \textcolor{red!70!black}{\textbf{0}};
Guilt \textcolor{red!70!black}{\textbf{0}};
Punishment \textcolor{red!70!black}{\textbf{0}};
Self-criticalness \textcolor{red!70!black}{\textbf{0}};
Agitation \textcolor{red!70!black}{\textbf{0}};
Irritability \textcolor{red!70!black}{\textbf{0}}
\\

\bottomrule
\end{tabularx}

\caption{
Example BDI-Sen annotations.
\textcolor{green!50!black}{\textbf{Green}} \textbf{1} denotes a positive symptom label and \textcolor{red!70!black}{\textbf{Red}} \textbf{0}
denotes a negative symptom label. The multi-positive example is one of the 144 sentences used in the multi-symptom retrieval evaluation in Section~\ref{sec:rq1_retrieval}. The third example contains depression-related language but no positive symptom label.
}
\label{tab:bdisen_examples}
\end{table*}

\begin{table}[t]
\centering
\small
\setlength{\tabcolsep}{4pt}
\renewcommand{\arraystretch}{1.08}

\begin{tabular}{lrrrr}
\toprule
\textbf{Setting} &
\textbf{$N$} &
\textbf{P} &
\textbf{R} &
\textbf{F1} \\
\midrule

\multicolumn{5}{l}{\textit{Gemma-3-4B-IT}} \\
Full                  & 2529 & 0.21 & 0.34 & 0.26 \\
At least one positive & 545  & 0.52 & 0.34 & 0.41 \\
Multi-positive        & 144  & 0.71 & 0.31 & 0.43 \\

\midrule

\multicolumn{5}{l}{\textit{Gemma-3-27B-IT}} \\
Full                  & 2529 & 0.28 & 0.41 & 0.33 \\
At least one positive & 545  & 0.55 & 0.41 & 0.47 \\
Multi-positive        & 144  & 0.67 & 0.39 & 0.49 \\

\bottomrule
\end{tabular}

\caption{
Zero-shot BDI-II symptom candidate extraction on BDI-Sen using the
item-name-plus-description symptom inventory at $\tau=0.70$.
\emph{Full} includes all 2,529 sentences,
\emph{At least one positive} includes the 545 sentences with at least
one mapped positive BDI-II item, and \emph{Multi-positive} includes the
144 sentences with at least two mapped positive BDI-II items.
}
\label{tab:bdisen_setting_results}
\end{table}

\section{Parameter Selection}
\label{app:parameters}

The main experiments use one frozen operating point, as described in
Section~\ref{sec:experimental_design}. This appendix records how those
values were selected.

\paragraph{Extraction confidence ($\tau$).}
Table~\ref{tab:subspan_threshold_sensitivity} shows the sensitivity of
MSR to the extraction confidence threshold on eRisk 2019. Increasing
$\tau$ reduces the number of retrieved posts and shifts the
precision--recall trade-off toward higher precision and lower recall.
We favor a moderate threshold rather than requiring high extractor
confidence, so that potentially relevant symptom evidence is retained
for downstream processing. We use $\tau=0.70$ for
Gemma-3-27B-IT and $\tau=0.75$ for Gemma-3-4B-IT
for a similar precision--recall balance.

\begin{figure*}[t]
    \centering
    \includegraphics[width=\textwidth]{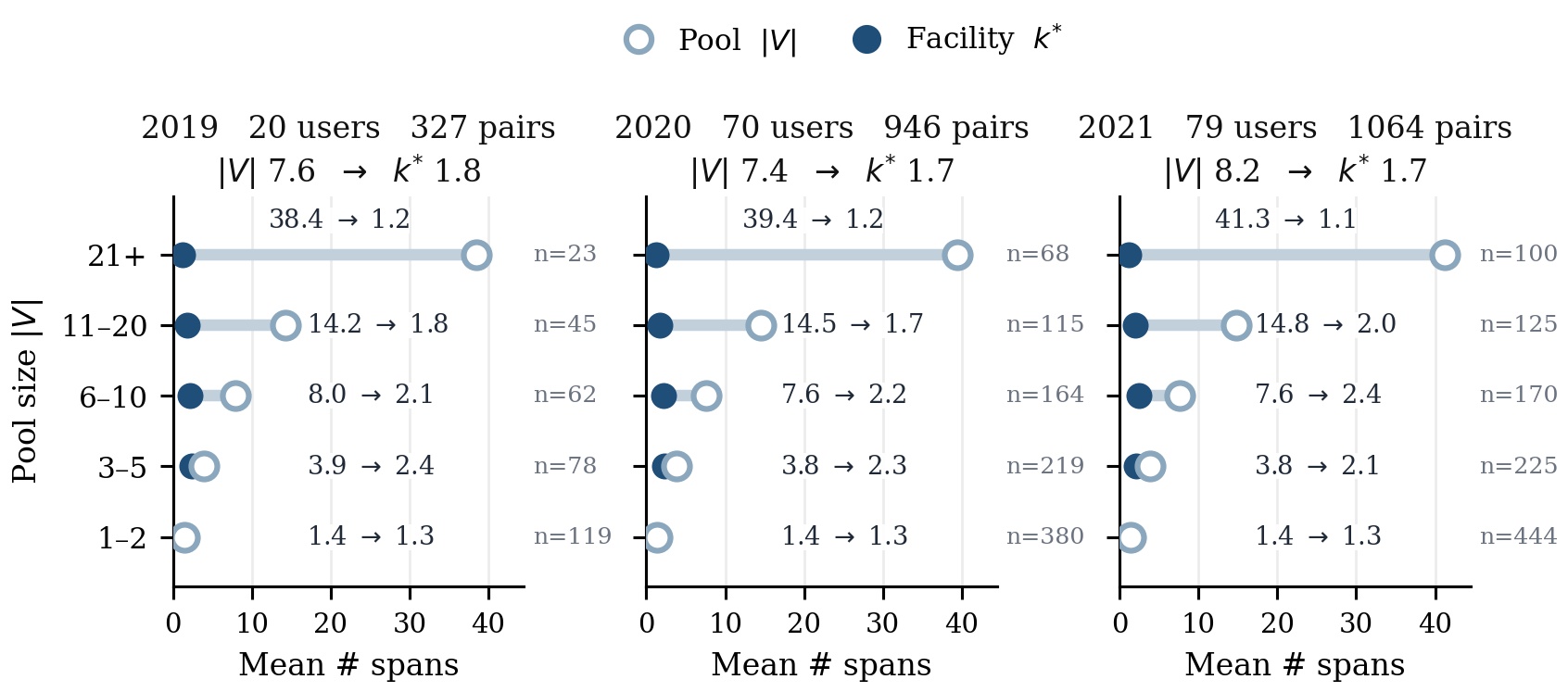}
    \caption{
    Facility Location behavior across eRisk 2019--2021. Each panel groups evidenced
    (user, item) pairs by their original candidate-pool size $|V|$.
    Open circles show the mean candidate-pool size and filled circles
    show the mean adaptively selected subset size $k^*$ within each
    pool-size group; $n$ denotes the number of evidenced pairs.
    Small pools remain largely unchanged, whereas larger pools are
    reduced to compact representative subsets.
    }
    \label{fig:pool_reduction}
\end{figure*}

\paragraph{Facility Location.}
Figure~\ref{fig:pool_reduction} summarizes the candidate evidence pools
and the subsets retained by the Facility Location-inspired submodular objective \cite{lin-bilmes-2011-class} for each evidenced
user--item pair. Across eRisk 2019, 2020, and 2021, the 327, 946, and
1,064 evidenced pairs have mean candidate-pool sizes $|V|$ of 7.6,
7.4, and 8.2 spans, respectively, with substantially larger pools in
the tail. We set $k_{\min}=1$ so that an evidenced pair always retains
at least one span, and $k_{\max}=20$ as a permissive upper bound to
prevent unusually large evidence pools from producing unnecessarily
long downstream prompts. 

The relevance weights in Eq.~\eqref{eq:relevance} and the stopping ratio
$\gamma$ used in the greedy optimization of
Eq.~\eqref{eq:marginal_gain} were selected from a predefined
36-configuration grid consisting of six weight combinations and
$\gamma\in\{0.05,0.075,0.10,0.125,0.15,0.20\}$.
Selection used user-level cross-validation on evidence-conditioned AHR,
with ACR and ADODL required not to decrease by more than one point
relative to the initial configuration $w_I=0.4$, $w_C=0.6$, and
$\gamma=0.10$. Protocol~A used eRisk 2019 and 2020 for parameter selection, with eRisk 2021 held out. Protocol~B used eRisk 2020 only, with both eRisk 2019 and 2021 held out. Both protocols selected $w_I=0.5$, $w_C=0.5$, and
$\gamma=0.20$, corresponding to the equal item--choice relevance weighting
in Eq.~\eqref{eq:relevance}. Extractor confidence enters the pipeline
through $\tau$ rather than the Facility relevance score. 

Under the selected
configuration, the mean subset sizes $k^*$ are 1.8, 1.7, and 1.7 for
2019, 2020, and 2021, respectively. As shown in
Figure~\ref{fig:pool_reduction}, pools containing only one or two spans
remain largely unchanged, whereas larger pools are reduced to compact
representative subsets. The selected subset sizes remain well below
$k_{\max}=20$, indicating that the retained evidence size is determined
primarily by the marginal gain and stopping rule in Eq.~\eqref{eq:marginal_gain} rather than by the upper bound.

\paragraph{Evidence-Gated Inference.}
The three-way gate is not grid-searched.
\textsc{Absent} is reserved for clear non-endorsement and is the only
label that forces a score of 0. \textsc{Uncertain} and \textsc{Present}
both proceed to ordinal inference.
We do not add a confidence threshold to the gate, avoiding a second
tuned cutoff on top of $\tau$.

\begin{table*}[t]
\centering
\footnotesize
\renewcommand{\arraystretch}{1.10}
\setlength{\tabcolsep}{12pt}

\begin{tabular}{
    @{}
    l
    c
    c
    c c c
    c c c
    @{}
}
\toprule

\multirow{2}{*}{\textbf{Model}}
&
\multirow{2}{*}{\makecell{\textbf{Confidence}\\\textbf{Threshold ($\tau$)}}}
&
\multirow{2}{*}{\makecell{\textbf{Avg. Posts}\\\textbf{per user}$\downarrow$}}
&
\multicolumn{3}{c}{\textbf{Macro}}
&
\multicolumn{3}{c}{\textbf{Micro}}
\\
\cmidrule(lr){4-6}
\cmidrule(l){7-9}
&
&
&
\textbf{P$\uparrow$}
&
\textbf{R$\uparrow$}
&
\textbf{F1$\uparrow$}
&
\textbf{P$\uparrow$}
&
\textbf{R$\uparrow$}
&
\textbf{F1$\uparrow$}
\\
\midrule
\multirow{4}{*}{\textbf{Gemma-3-4B-IT}}
& 0.70 & 142.10 & 0.13 & \textbf{0.76} & 0.22 & 0.12 & \textbf{0.80} & 0.21 \\
& 0.75 & 121.80 & \underline{0.15} & \underline{0.73} & \underline{0.24} & \underline{0.14} & \underline{0.76} & \underline{0.23} \\
& 0.80 & \underline{121.60} & 0.15 & 0.73 & 0.24 & 0.14 & 0.76 & 0.23 \\
& 0.85 & \textbf{73.40} & \textbf{0.18} & 0.56 & \textbf{0.27} & \textbf{0.18} & 0.61 & \textbf{0.28} \\
\midrule
\multirow{4}{*}{\textbf{Gemma-3-27B-IT}}
& 0.70 & 86.00 & 0.17 & \textbf{0.68} & 0.27 & 0.17 & \textbf{0.67} & 0.27 \\
& 0.75 & 47.50 & 0.23 & \underline{0.55} & \underline{0.32} & 0.24 & \underline{0.52} & \textbf{0.33} \\
& 0.80 & \underline{43.90} & \underline{0.25} & 0.53 & \textbf{0.33} & \underline{0.25} & 0.50 & \underline{0.33} \\
& 0.85 & \textbf{25.60} & \textbf{0.30} & 0.33 & 0.30 & \textbf{0.29} & 0.34 & 0.31 \\
\bottomrule
\end{tabular}
\caption{
Sensitivity of MSR to the extraction confidence
threshold on eRisk 2019, using the same BDI-Sen TITLE+TEXT gold set
(441 posts) as Table~\ref{tab:erisk_bdisen_retrieval_main}.
Raising the threshold reduces retrieved posts and trades recall for
precision. Within each model, \textbf{bold} is best and
\underline{underline} is second-best on the unrounded scores
(lower is better for average posts per user).
The main-table operating points are 4B at $\tau=0.75$ and 27B at $\tau=0.70$.
}
\label{tab:subspan_threshold_sensitivity}
\end{table*}

\section{Full-Questionnaire Evaluation}
\label{app:full21}

Social-media histories do not provide complete observations of every
BDI-II symptom. A missing retrieved span may reflect either symptom absence or insufficient evidence in the user's posts. We therefore use evidence-conditioned inference as our primary setting, where a BDI-II item is scored only when grounded evidence is available and unsupported items remain unobserved. To retain comparability with prior work, we additionally report a
full-21 setting in which all 21 items are evaluated and items without
grounded evidence are assigned a score of 0. The baseline results in
this setting are our Gemma-3-27B reimplementations rather than values
reported in the original publications. Let $E_i$ denote the set of evaluated BDI-II items for user $u_i$, and
let $m_i=|E_i|$. The standard eRisk metrics assume predictions for all
21 BDI-II items. In the evidence-conditioned setting, AHR and ACR are
computed only over $E_i$, while ADODL uses $3m_i$ as the maximum
possible aggregate score difference rather than $63$. The final metrics are obtained by averaging the corresponding per-user
scores. For the standard setting, the average is taken over all users.
For the evidence-conditioned setting, we average over
$U'=\{u_i:m_i>0\}$:

Table~\ref{tab:metric_adaptation} summarizes these adaptations.
DCHR is reported only for the full-21 setting because assigning a
depression category requires a complete questionnaire-level score. Here, $C(\cdot)$ maps the total BDI-II score to the corresponding
severity category: minimal (0--9), mild (10--18), moderate (19--29),
or severe (30--63). DCHR has no evidence-conditioned counterpart because $E_i$ does not
define a complete questionnaire total.

\begin{table*}[t]
\centering
\small
\begin{tabular}{lcc}
\toprule
\textbf{Metric} &
\textbf{Full 21-item evaluation} &
\textbf{Evidence-conditioned evaluation} \\
\midrule

\makecell[l]{Average Hit Rate\\(AHR)} &
$\displaystyle
\frac{1}{N}
\sum_{i=1}^{N}
\frac{1}{21}
\sum_{k=1}^{21}
\mathbb{I}[\hat y_i^k=y_i^k]
$
&
$\displaystyle
\frac{1}{|U'|}
\sum_{u_i\in U'}
\frac{1}{m_i}
\sum_{k\in E_i}
\mathbb{I}[\hat y_i^k=y_i^k]
$
\\[5mm]

\makecell[l]{Average Closeness Rate\\(ACR)} &
$\displaystyle
\frac{1}{N}
\sum_{i=1}^{N}
\frac{1}{21}
\sum_{k=1}^{21}
\frac{3-|\hat y_i^k-y_i^k|}{3}
$
&
$\displaystyle
\frac{1}{|U'|}
\sum_{u_i\in U'}
\frac{1}{m_i}
\sum_{k\in E_i}
\frac{3-|\hat y_i^k-y_i^k|}{3}
$
\\[5mm]

\makecell[l]{Average Difference between\\Overall Depression Levels\\(ADODL)} &
$\displaystyle
\frac{1}{N}
\sum_{i=1}^{N}
\frac{
63-\left|
\sum_{k=1}^{21}\hat y_i^k-
\sum_{k=1}^{21}y_i^k
\right|
}{63}
$
&
$\displaystyle
\frac{1}{|U'|}
\sum_{u_i\in U'}
\frac{
3m_i-\left|
\sum_{k\in E_i}\hat y_i^k-
\sum_{k\in E_i}y_i^k
\right|
}{3m_i}
$
\\[5mm]

\makecell[l]{Depression Category Hit Rate\\(DCHR)} &
$\displaystyle
\frac{1}{N}
\sum_{i=1}^{N}
\mathbb{I}\!\left[
C\!\left(\sum_{k=1}^{21}\hat y_i^k\right)
=
C\!\left(\sum_{k=1}^{21}y_i^k\right)
\right]
$
&
-- \\

\bottomrule
\end{tabular}

\caption{
Standard eRisk metrics and their evidence-conditioned adaptations.
Here, $E_i$ denotes the set of evaluated BDI-II items for user $u_i$,
$m_i=|E_i|$, $U'=\{u_i:m_i>0\}$, and $N$ is the total number of users.
In the evidence-conditioned setting, AHR and ACR are computed over the
available items in $E_i$, while ADODL uses $3m_i$ rather than $63$ as
the maximum aggregate score difference. DCHR is defined only for the
complete 21-item questionnaire.
}
\label{tab:metric_adaptation}
\end{table*}

\begin{table}[t]
\centering
\footnotesize
\setlength{\tabcolsep}{2.8pt}
\renewcommand{\arraystretch}{0.95}

\begin{tabular}{@{}lcccc@{}}
\toprule
\textbf{Method}
& \textbf{Cov.}
& \textbf{P$\uparrow$}
& \textbf{R$\uparrow$}
& \textbf{F1$\uparrow$} \\
\midrule

\multicolumn{5}{c}{\textbf{Matched budget ($K=2$)}} \\
\midrule

all-mpnet-base-v2$^{\textcolor{blue}{a,c}}$
& 0.42 & 0.56 & 0.38 & 0.45 \\

all-MiniLM-L12-v2$^{\textcolor{blue}{b}}$
& 0.42 & 0.56 & 0.38 & 0.45 \\

msmarco-MiniLM-L12-v3$^{\textcolor{blue}{c}}$
& 0.39 & 0.54 & 0.37 & 0.44 \\

msmarco-distilbert-base-v4$^{\textcolor{blue}{c}}$
& 0.40 & 0.55 & 0.37 & 0.44 \\

\rowcolor{oursrow}
\textbf{MSR (Gemma-3-27B-IT)}
& 0.39 & \textbf{0.68} & \textbf{0.40} & \textbf{0.50} \\

\midrule
\multicolumn{5}{c}{\textbf{$\sim$2$\times$ budget ($K=4$)}} \\
\midrule

all-mpnet-base-v2$^{\textcolor{blue}{a,c}}$
& 0.60 & 0.42 & 0.56 & 0.48 \\

all-MiniLM-L12-v2$^{\textcolor{blue}{b}}$
& 0.62 & 0.42 & \textbf{0.57} & 0.49 \\

msmarco-MiniLM-L12-v3$^{\textcolor{blue}{c}}$
& 0.55 & 0.39 & 0.53 & 0.45 \\

msmarco-distilbert-base-v4$^{\textcolor{blue}{c}}$
& 0.58 & 0.41 & 0.56 & 0.47 \\

\rowcolor{oursrow}
\textbf{MSR (Gemma-3-27B-IT)}
& 0.43 & \textbf{0.68} & 0.40 & \textbf{0.50} \\

\bottomrule
\end{tabular}

\vspace{2pt}

\parbox{\columnwidth}{
\scriptsize
$^{a}$~APB~\cite{bascunana2024apb};
$^{b}$~Two-Hop~\cite{xu2026social};
$^{c}$~Adaptive RAG~\cite{ravenda2025llms}.
}

\caption{
Multi-symptom BDI-II item retrieval on 144 BDI-Sen examples. The
matched-budget ($K=2$) results are repeated for reference alongside an
approximately $2\times$ dense-budget setting ($K=4$). Best results
within each setting are boldfaced.
}
\label{tab:multisymptom_app}
\end{table}

Table~\ref{tab:full21_results} shows how SREE behaves when items
without grounded evidence are assigned a score of 0. SREE achieves the
highest AHR on eRisk 2019 and 2020 and the highest ADODL on eRisk 2020,
while its performance on the remaining metrics is generally below the
strongest competing baselines, particularly on eRisk 2021. This
contrasts with the evidence-conditioned evaluation, where items without
grounded evidence remain unobserved rather than being interpreted as
symptom absence. The comparison highlights the effect of the evaluation setting:
full-21 reconstruction requires predictions for every BDI-II item,
whereas evidence-conditioned evaluation scores only items with grounded
textual evidence.

\begin{table*}[t]
\centering
\footnotesize
\renewcommand{\arraystretch}{1}
\setlength{\tabcolsep}{4pt}

\begin{tabularx}{0.98\textwidth}{
    @{}
    >{\raggedright\arraybackslash}p{1.8cm}
    >{\raggedright\arraybackslash}X
    >{\centering\arraybackslash}p{1.2cm}
    >{\centering\arraybackslash}p{1.2cm}
    >{\centering\arraybackslash}p{1.5cm}
    >{\centering\arraybackslash}p{1.3cm}
    @{}
}
\toprule

\multirow{2}{*}{\textbf{Collection}}
&
\multirow{2}{*}{\textbf{Methods}}
&
\multicolumn{2}{c}{\textbf{Symptom Metrics}}
&
\multicolumn{2}{c}{\textbf{Questionnaire Metrics}}
\\

\cmidrule(lr){3-4}
\cmidrule(l){5-6}

&
&
\textbf{AHR$\uparrow$}
&
\textbf{ACR$\uparrow$}
&
\textbf{ADODL$\uparrow$}
&
\textbf{DCHR$\uparrow$}
\\

\midrule


\multirow{3}{*}{
\makecell[l]{\textbf{eRisk 2019}\\
20 users \\ 420 items}
}
&
BDI-biased Summarization~\cite{aragon2024delving}
&
\underline{32.38}
&
61.11
&
72.54
&
15.00
\\

&
Fixed Top-$k$ Retrieval~\cite{wang2024explainable}
&
26.90
&
\underline{69.60}
&
\underline{79.29}
&
15.00
\\

&
Adaptive RAG (aRAG)~\cite{ravenda2025llms}
&
31.19
&
\textbf{70.32}
&
\textbf{82.86}
&
\textbf{35.00}
\\

\noalign{\vskip 1.5pt}
\cdashline{2-6}
\noalign{\vskip 3.5pt}

\rowcolor{oursrow}
\textbf{Ours}
&
\makecell[l]{\textbf{SREE Framework}\\
(MSR + Evidence Selection + Evidence-Gated Inference)}
&
\makecell{\textbf{35.71}\\\posdelta{3.33}}
&
\makecell{67.70\\\negdelta{2.62}}
&
\makecell{76.59\\\negdelta{6.27}}
&
\makecell{\underline{20.00}\\\negdelta{15.00}}
\\

\midrule


\multirow{3}{*}{
\makecell[l]{\textbf{eRisk 2020}\\
70 users \\ 1470 items}
}
&
BDI-biased Summarization~\cite{aragon2024delving}
&
\underline{33.81}
&
62.72
&
73.29
&
20.00
\\

&
Fixed Top-$k$ Retrieval~\cite{wang2024explainable}
&
30.61
&
\textbf{71.27}
&
\underline{81.70}
&
\underline{34.29}
\\

&
Adaptive RAG (aRAG)~\cite{ravenda2025llms}
&
30.68
&
\underline{69.89}
&
80.82
&
\textbf{37.14}
\\

\noalign{\vskip 1.5pt}
\cdashline{2-6}
\noalign{\vskip 3.5pt}

\rowcolor{oursrow}
\textbf{Ours}
&
\makecell[l]{\textbf{SREE Framework}\\
(MSR + Evidence Selection + Evidence-Gated Inference)}
&
\makecell{\textbf{36.33}\\\posdelta{2.52}}
&
\makecell{69.07\\\negdelta{2.20}}
&
\makecell{\textbf{82.22}\\\posdelta{0.52}}
&
\makecell{\underline{34.29}\\\negdelta{2.85}}
\\

\midrule


\multirow{4}{*}{
\makecell[l]{\textbf{eRisk 2021}\\
79 users \\ 1659 items}
}
&
BDI-biased Summarization~\cite{aragon2024delving}
&
23.33
&
55.98
&
65.14
&
12.66
\\

&
Fixed Top-$k$ Retrieval~\cite{wang2024explainable}
&
\textbf{34.30}
&
\textbf{72.27}
&
\textbf{85.09}
&
\textbf{55.70}
\\

&
Adaptive RAG (aRAG)~\cite{ravenda2025llms}
&
\underline{33.76}
&
\underline{70.81}
&
\underline{82.62}
&
\underline{51.90}
\\

&
AdaDepression~\cite{xu2026social}
&
30.38
&
68.84
&
79.85
&
50.63
\\

\noalign{\vskip 1.5pt}
\cdashline{2-6}
\noalign{\vskip 3.5pt}

\rowcolor{oursrow}
\textbf{Ours}
&
\makecell[l]{\textbf{SREE Framework}\\
(MSR + Evidence Selection + Evidence-Gated Inference)}
&
\makecell{30.74\\\negdelta{3.56}}
&
\makecell{65.38\\\negdelta{6.89}}
&
\makecell{76.91\\\negdelta{8.18}}
&
\makecell{20.25\\\negdelta{35.45}}
\\

\bottomrule
\end{tabularx}

\caption{
Full-21 BDI-II inference results on the eRisk collections (in \%),
where all 21 items are evaluated and items without grounded evidence
are assigned a score of 0.
\textbf{Bold} indicates the best result and \underline{underline}
indicates the second-best result within each collection.
For \textbf{SREE}, the value shown below each score denotes the
difference from the best competing baseline for that metric,
$\Delta=\text{SREE}-\text{Best Baseline}$.
Higher values are better.
}
\label{tab:full21_results}

\end{table*}

\begin{table*}[t]
\centering
\footnotesize
\renewcommand{\arraystretch}{0.9}

\setlength{\tabcolsep}{9.4pt}

\begin{tabular}{
    @{}
    l
    c
    c c c
    c c c
    @{}
}
\toprule

\multirow{2}{*}{\textbf{Method}}
&
\multirow{2}{*}{\textbf{Avg Posts $\downarrow$}}
&
\multicolumn{3}{c}{\textbf{Macro}}
&
\multicolumn{3}{c}{\textbf{Micro}}
\\

\cmidrule(lr){3-5}
\cmidrule(l){6-8}

&
&
\textbf{P$\uparrow$}
&
\textbf{R$\uparrow$}
&
\textbf{F1$\uparrow$}
&
\textbf{P$\uparrow$}
&
\textbf{R$\uparrow$}
&
\textbf{F1$\uparrow$}
\\

\midrule

Fixed Top-$k$ Retrieval~\cite{wang2024explainable}
&
147.45
&
0.09
&
0.69
&
0.16
&
0.09
&
0.61
&
0.16
\\

Adaptive-$k^{\star}$ Retrieval~\cite{ravenda2025llms}
&
123.45
&
0.10
&
0.65
&
0.16
&
0.09
&
0.52
&
0.16
\\

Two-Hop Retrieval~\cite{xu2026social}
&
142.60
&
0.10
&
\underline{0.73}
&
0.17
&
0.10
&
0.67
&
0.18
\\

\noalign{\vskip 1.5pt}
\cmidrule(lr){1-8}
\noalign{\vskip 3.5pt}

\rowcolor{oursrow}
\shortstack[l]{
\textbf{Multi-Symptom Span Retrieval (MSR)}\\
\textcolor{blue}{(Gemma-3-4B-IT, $\tau=0.75$)}
}
&
\underline{121.80}
&
\underline{0.15}
&
\textbf{0.73}
&
\underline{0.24}
&
\underline{0.14}
&
\textbf{0.76}
&
\underline{0.23}
\\

\noalign{\vskip 2.5pt}

\rowcolor{oursrow}
\shortstack[l]{
\textbf{Multi-Symptom Span Retrieval (MSR)}\\
\textcolor{blue}{(Gemma-3-27B-IT, $\tau=0.70$)}
}
&
\textbf{86.00}
&
\textbf{0.17}
&
0.68
&
\textbf{0.27}
&
\textbf{0.17}
&
\underline{0.67}
&
\textbf{0.27}
\\

\bottomrule
\end{tabular}


\caption{
Post-level evidence retrieval on eRisk 2019 using BDI-Sen annotations.
A post is treated as gold-positive if its \textsc{Title+Text} contains at
least one BDI-Sen sentence with a positive BDI-II symptom annotation,
yielding 441 gold-positive posts across 20 users. Avg.\ Posts reports the
mean number of retrieved posts per user. Macro P/R/F1 are computed per
user and then averaged, whereas micro P/R/F1 pool all retrieved and
gold-positive posts across users. Fixed Top-$k$, aRAG, and Two-Hop use
\texttt{all-MiniLM-L12-v2} as the common dense retrieval backbone.
Shaded rows show MSR with Gemma-3-4B-IT ($\tau=0.75$) and
Gemma-3-27B-IT ($\tau=0.70$). \textbf{Bold} and
\underline{underline} denote the best and second-best results based on
unrounded values. Higher P/R/F1 and lower Avg.\ Posts are better.
}
\label{tab:erisk_bdisen_retrieval_appendix}

\end{table*}

\section{Additional Evidence Retrieval Analysis}
\label{app:retrieval_analysis}

\noindent \textbf{Post-Level Retrieval Diagnostics} Table~\ref{tab:erisk_bdisen_retrieval_appendix} reports the complete
post-level retrieval results on eRisk 2019, including both macro- and
micro-averaged metrics. BDI-Sen annotations identify 441 gold-positive
posts. At the downstream operating point, MSR with Gemma-3-27B-IT
retains an average of 86 posts per user, compared with 123.45 for aRAG.
MSR retains 297 gold-positive posts, whereas aRAG retains 230, despite
using a smaller retrieval budget. aRAG additionally retrieves 2,239
false-positive posts, 1,663 of which MSR excludes.

The difference reflects how evidence is admitted by the two retrieval
strategies. Dense retrieval ranks posts by semantic proximity to BDI-II
response text, so lexical or topical overlap can surface advice,
well-wishing, benign self-reports, or other symptom-related discussion
without attributable evidence for the item. MSR instead requires a
grounded first-person symptom span before a post enters the evidence
pool. Table~\ref{tab:retrieval_examples} provides representative examples
of both failure modes: semantically related but unsupported posts retrieved
by dense methods, and attributable gold-positive posts recovered by MSR
but missed by dense retrieval.

\begin{table*}[t]
\centering
\footnotesize
\renewcommand{\arraystretch}{1.10}
\setlength{\tabcolsep}{5pt}

\begin{tabularx}{0.98\textwidth}{
    @{}
    >{\raggedright\arraybackslash}X
    >{\raggedright\arraybackslash}p{4.2cm}
    >{\raggedright\arraybackslash}p{4.2cm}
    @{}
}
\toprule

\textbf{Paraphrased posts}
&
\shortstack[l]{\textbf{Dense Retrieval}\\
\textbf{(e.g., aRAG Framework)}}
&
\shortstack[l]{\textbf{Grounded Span Retrieval}\\
\textbf{(e.g., SREE Framework)}}
\\
\midrule

``You are \traptext{not worthless}. You are a thoughtful,
caring person and deserve a good life. Stay strong---I'm here for you.''
&
\badoutcome{Retrieved}.
Nearest BDI-II option:
\emph{I do not feel I am worthless.}
&
\goodoutcome{Rejected}.
No grounded first-person symptom span.
\\

\midrule

``\traptext{Alcohol may be contributing to your problems},
and \traptext{you should consider seeing another psychologist}.''
&
\badoutcome{Retrieved}.
Nearest to Loss-of-Interest /
depression-related response language.
&
\goodoutcome{Rejected}.
The post describes another person's condition rather than
the user's own symptoms.
\\

\midrule

``\traptext{Antidepressants can be unpleasant, but research suggests
they may help}. I hope you feel better.''
&
\badoutcome{Retrieved}.
Nearest BDI-II option:
\emph{I am so restless or agitated that it is hard to stay still.}
&
\goodoutcome{Rejected}.
No grounded first-person agitation span.
\\

\midrule

``The user looks forward to a \traptext{future relationship}
and enjoys sentimental romance movies.''
&
\badoutcome{Retrieved}.
Nearest BDI-II option:
\emph{I feel my future is hopeless and will only get worse.}
&
\goodoutcome{Rejected}.
No grounded first-person pessimism span.
\\

\midrule

``You're human and allowed to make mistakes.
You are \traptext{not worthless or a failure}.
I'm here for you.''
&
\badoutcome{Retrieved}.
Nearest BDI-II option:
\emph{I feel I am a total failure as a person.}
&
\goodoutcome{Rejected}.
No grounded first-person symptom span.
\\

\midrule

``\groundtext{I hate myself every day because of my grades}.
Anything short of perfect leaves me disappointed and crying.''
&
\badoutcome{Missed}.
The gold post is not surfaced.
&
\goodoutcome{Retrieved}.
Grounded first-person span supports
\emph{Self-dislike}.
\\

\midrule

``My depression shows in my face---I'm bad at hiding my feelings,
\groundtext{which is something I hate about myself}.
I go quiet and my head is in turmoil.''
&
\badoutcome{Missed}.
The gold post is not surfaced.
&
\goodoutcome{Retrieved}.
Grounded first-person span supports
\emph{Self-dislike}.
\\

\midrule

``\groundtext{I have dealt with depression since middle school}.
Some days are especially difficult, and
\groundtext{life can feel hopeless}.''
&
\badoutcome{Missed}.
The gold post is not surfaced.
&
\goodoutcome{Retrieved}.
Grounded first-person spans support
\emph{Sadness} and \emph{Pessimism}.
\\

\bottomrule
\end{tabularx}

\caption{
Illustrative retrieval errors from eRisk 2019 contrasting dense semantic
retrieval with grounded sub-span retrieval.
\textcolor{crossorange}{Orange} marks misleading lexical or semantic cues,
while \textcolor{deltapos}{green} marks attributable first-person symptom
evidence. \textcolor{red}{Red} outcomes indicate false retrievals or misses. \textcolor{deltapos}{Green} outcomes
indicate correct rejection or retrieval.
}
\label{tab:retrieval_examples}
\end{table*}

\noindent \textbf{Macro--micro behavior.} Macro averaging weights users equally, whereas micro averaging gives greater influence to users with more gold-positive posts. For aRAG,
recall drops from 0.65 (macro) to 0.52 (micro), consistent with lower
recall for users contributing more gold-positive posts. MSR shows nearly
identical macro and micro recall (0.68 vs.\ 0.67), indicating a more
stable retrieval profile across aggregation schemes. We treat this as a
secondary diagnostic rather than  a primary result.

MSR is not error-free: the extractor can still admit imperfectly
attributed spans, including occasional second-person statements. The key
distinction is therefore the evidence-admission criterion: MSR requires
an explicit grounded span, whereas dense retrieval relies on semantic
proximity alone.
\newline


\noindent \textbf{Multi-Symptom Retrieval under Expanded Budget.}
Table~\ref{tab:multisymptom_app} extends the matched-budget analysis in
Section~\ref{sec:rq1_retrieval} by doubling the dense retrieval budget
from $K=2$ to $K=4$. Increasing the budget improves dense coverage and
recall, but reduces precision. For example, all-MiniLM-L12-v2 increases
recall from 0.38 to 0.57 and coverage from 0.42 to 0.62, while precision
drops from 0.56 to 0.42. Its F1 rises to 0.49 but remains below MSR
(0.50), despite emitting four candidates per example compared with
MSR's natural prediction set (mean 1.74). The expanded-budget results thus reinforce the $K=2$ finding: dense retrieval can recover
additional symptoms by admitting more candidates, but this also
introduces more unsupported labels. MSR maintains higher precision and
F1 with a much smaller grounded set.

\section{Prompt Templates}
\label{app:prompts}

This section provides the prompt templates used throughout SREE.
Figure~\ref{fig:prompt} shows the zero-shot MSR prompt used to identify
symptom-specific candidate spans from individual posts, with the BDI-II
symptom inventory used in the prompt shown in Figure~\ref{fig:items_desc}.
Figures~\ref{fig:evidence_gate_prompt} and~\ref{fig:ordinal_prompt}
show the Evidence-Gated Inference prompt and the subsequent ordinal
inference prompt used to assign the 0--3 BDI-II response.

\begin{figure*}[t]
\centering

\begin{tcolorbox}[
    colback=gray!5!white,
    colframe=black!75!black,
    boxrule=0.5pt,
    arc=2pt,
    left=5pt,
    right=5pt,
    top=5pt,
    bottom=5pt,
    width=\textwidth,
    title=\textbf{BDI-II Symptom Inventory -- \texttt{\{items\_desc\}}},
    fonttitle=\normalsize
]
\setlength{\parindent}{0pt}
\setlength{\parskip}{0pt}

\normalsize
\raggedright


\medskip

1. Sadness --- how sad the user feels\\
2. Pessimism --- how discouraged the user is about future\\
3. Past Failure --- how much the user feels like a failure\\
4. Loss of Pleasure --- how much the user loses pleasure from things\\
5. Guilty Feelings --- how often the user feels guilty\\
6. Punishment Feelings --- how much the user feels punished\\
7. Self-Dislike --- how much the user feels disappointed about him/herself\\
8. Self-Criticalness --- how often the user criticizes or blames him/herself\\
9. Suicidal Thoughts or Wishes --- how much the user thinks about killing him/herself\\
10. Crying --- how often the user cries\\
11. Agitation --- how much the user feels restless or agitated\\
12. Loss of Interest --- how much the user loses interest in things\\
13. Indecisiveness --- how difficult the user to make decisions\\
14. Worthlessness --- how much the user feels worthless\\
15. Loss of Energy --- how much the user loses energy\\
16. Changes in Sleeping Pattern --- how much the user experienced changes in sleeping\\
17. Irritability --- how much the user feels irritable\\
18. Changes in Appetite --- how much the user experienced changes in appetite\\
19. Concentration Difficulty --- how difficult the user to concentrate\\
20. Tiredness or Fatigue --- how much the user feels tired or fatigued\\
21. Loss of Interest in Sex --- how much the user loses interest in sex

\end{tcolorbox}

\caption{BDI-II symptom inventory used in the MSR prompt.}
\label{fig:items_desc}
\end{figure*}

\begin{figure*}[t]
\centering

\begin{tcolorbox}[
    colback=gray!5!white,
    colframe=black!75!black,
    boxrule=0.5pt,
    arc=2pt,
    left=5pt,
    right=5pt,
    top=3pt,
    bottom=3pt,
    width=\textwidth,
    title=\textbf{Zero-shot MSR Prompt},
    fonttitle=\normalsize
]
\normalsize
\raggedright

You are a clinical psychologist assistant.\\

Your task: Given one social media post written by a user, evaluate all 21 BDI-II symptoms and detect every symptom that is explicitly expressed as the user's own self-reported symptom evidence in the post. A symptom is relevant only when the post expresses the user's own mental state, belief, behavior, or physical condition. The evidence must be explicit or strongly implied by the post itself. You MUST check each symptom independently and systematically. You MUST evaluate all 21 symptoms, but you MUST NOT invent symptoms without textual evidence.\\

\smallskip
For each BDI-II symptom, do the following:

Step 1: Decide whether the symptom is clearly expressed as the user's own symptom.\\
Step 2: If NOT expressed, do NOT output it.\\
Step 3: If expressed, you MUST output:

\hspace*{1em}-- item\_id: integer from 1 to 21\\
\hspace*{1em}-- item\_name: exact BDI-II item name\\
\hspace*{1em}-- span: the shortest exact substring from the post that supports the symptom\\
\hspace*{1em}-- confidence: a float from 0.0 to 1.0 indicating how strongly the text supports the symptom\\

\smallskip
Do NOT mark a symptom as relevant when the post is a general statement about depression, symptoms, or mental health; the post describes another person's symptom; the post is advice or encouragement directed at someone else; the post is hypothetical, educational, quoted, or unrelated; the post merely contains a symptom-related word without self-reported evidence; the post does not express the user's own symptom evidence; or the post is a control/non-depression post.

\smallskip
If no user-expressed BDI-II symptom is present, output:
\{\{"candidates": []\}\}

\smallskip
BDI-II symptom inventory: \{items\_desc\}

\smallskip
Post: """\{post\}"""

\smallskip
Output exactly one JSON object using this schema:

\{\{\\
\hspace*{1em}"candidates": [\\
\hspace*{2em}\{\{\\
\hspace*{3em}"item\_id": $<$int$>$,\\
\hspace*{3em}"item\_name": "$<$string$>$",\\
\hspace*{3em}"span": "$<$string$>$",\\
\hspace*{3em}"confidence": $<$float$>$\\
\hspace*{2em}\}\}\\
\hspace*{1em}]\\
\}\}

\smallskip
Critical output rules: \\
Output only valid JSON without explanations, reasoning, markdown, or code fences. Do not repeat, rewrite, or summarize the post. The output must start with `\{\{' and end with `\}\}'.

\end{tcolorbox}

\caption{Zero-shot MSR prompt for BDI-II symptom candidate extraction.
\texttt{\{items\_desc\}} refers to the inventory in Figure~\ref{fig:items_desc};
\texttt{\{post\}} is replaced with the input post.}
\label{fig:prompt}
\end{figure*}

\begin{figure*}[t]
\centering

\begin{tcolorbox}[
    colback=gray!5!white,
    colframe=black!75!black,
    boxrule=0.5pt,
    arc=2pt,
    left=5pt,
    right=5pt,
    top=3pt,
    bottom=3pt,
    width=\textwidth,
    title=\textbf{Ordinal Inference Prompt},
    fonttitle=\normalsize
]
\normalsize
\raggedright

You are a clinical psychologist assistant.

Your task: Given evidence taken from one user's own social media posts, rate a single BDI-II symptom on its 0--3 severity scale, using ONLY the user's own self-reported state.

\smallskip
You are evaluating exactly ONE symptom:

BDI-II Item \{item\_id\} --- \{item\_name\} (\{query\}).

\smallskip
The evidence below is grouped by post: each of the user's posts is shown in full, followed by the sub-spans identified within it. Treat the posts and their sub-spans as the user speaking in the first person about their own mental state, belief, behavior, or physical condition. Use the full posts as context and the sub-spans as the focus.

\smallskip
\textbf{Evidence:}

\{evidence\}

\smallskip
\textbf{Severity scale for this symptom (choose exactly one level, 0 to 3):}

\{scoring\_block\}

\smallskip
You MUST follow these rules: Rate ONLY this symptom, based on the user's own self-reported evidence. Choose the highest severity level that is clearly and repeatedly supported across the evidence; do NOT over-rate a single weak, vague, or ambiguous mention. Ignore spans that describe another person, are advice to others, are hypothetical, quoted, or educational, or that do not actually express this symptom. If the evidence does not genuinely support this symptom as the user's own, choose 0. Also report your confidence in the chosen severity as a float from 0.0 to 1.0.

\smallskip
\textbf{Critical output rules:}

Output EXACTLY one JSON object:
\{\{"score": $<$0 | 1 | 2 | 3$>$, "confidence": $<$float 0.0--1.0$>$\}\}.

``score'' is the chosen severity level; ``confidence'' is how confident you are in that choice. Do NOT output any other text, explanation, reasoning, markdown, or code fences. The output must start with `\{' and end with `\}'.

\smallskip
\textbf{Answer:}

\end{tcolorbox}

\caption{Zero-shot ordinal inference prompt used to assign a 0--3 BDI-II response from the selected evidence.}
\label{fig:ordinal_prompt}
\end{figure*}

\begin{figure*}[t]
\centering

\begin{tcolorbox}[
    colback=gray!5!white,
    colframe=black!75!black,
    boxrule=0.5pt,
    arc=2pt,
    left=5pt,
    right=5pt,
    top=3pt,
    bottom=3pt,
    width=\textwidth,
    title=\textbf{Evidence-Gated Inference Prompt},
    fonttitle=\normalsize
]
\normalsize
\raggedright

You are a clinical psychologist assistant.

\smallskip
Your task: Given evidence taken from one user's own social media posts, decide occupancy for this specific BDI-II symptom. Do NOT rate severity. Do NOT choose 0, 1, 2, or 3.

\smallskip
You are evaluating exactly ONE symptom:

BDI-II Item \{item\_id\} --- \{item\_name\} (\{query\}).

\smallskip
The evidence below is grouped by post: each of the user's posts is shown in full, followed by the sub-spans identified within it.

\smallskip
This is a high-precision rejector, not a full occupancy classifier. ABSENT is the only decision that later forces a score of 0, so it must be reserved for cases you are confident are not endorsements. If you are not sure, choose UNCERTAIN. Do not force a binary PRESENT/ABSENT call.

\smallskip
Choose exactly one of:

\textbf{ABSENT:} strong exclusion evidence that this is NOT a first-person endorsement of this item. Typical exclusions: another person's symptom; second-person or general advice; quoted content; entertainment or media (TV/movie/fictional characters); hypothetical; metaphor or rhetoric with no literal self-report; educational or general discussion; clearly a different BDI-II item. Use ABSENT only when the exclusion is clear.

\textbf{PRESENT:} clear first-person evidence that the user experiences this target symptom. Mild or occasional language still counts as PRESENT if it is genuinely self-reported and item-relevant.

\textbf{UNCERTAIN:} mild, indirect, abstract, mixed, or ambiguous evidence. This includes cases that might be a weak self-report of this symptom, might be a different item, or might be exclusion-like but are not clear enough to reject. Cognitive items (worthlessness, past failure, self-dislike, punishment feelings, indecisiveness) are often indirect: if the first-person meaning is possible but not certain, choose UNCERTAIN, not ABSENT.

\smallskip
Asymmetric rule: when in doubt between ABSENT and UNCERTAIN, choose UNCERTAIN. When in doubt between PRESENT and UNCERTAIN, choose UNCERTAIN. Never choose ABSENT just because the evidence is weak.

\smallskip
You MUST follow these rules: Decide ONLY occupancy for this symptom. Do not infer occupancy from the number of posts or spans. Do not consider how severe the symptom would be if present. Also report your confidence in the occupancy decision as a float from 0.0 to 1.0.

\smallskip
\textbf{Critical output rules:}

Output EXACTLY one JSON object:

\{\{"occupancy": "ABSENT"|"UNCERTAIN"|"PRESENT", "confidence": $<$float 0.0--1.0$>$\}\}.

Do NOT output any other text, explanation, reasoning, markdown, or code fences. The output must start with `\{' and end with `\}'.

\smallskip
\textbf{Evidence:}

\{evidence\}

\smallskip
\textbf{Answer:}

\end{tcolorbox}

\caption{Zero-shot evidence-gated inference prompt used to determine whether the selected evidence provides attributable support for the target BDI-II item.}
\label{fig:evidence_gate_prompt}
\end{figure*}

\end{document}